\documentclass[
]{ceurart}

\usepackage{listings}
\usepackage[T1]{fontenc}
\usepackage{graphicx}
\usepackage{subfigure}
\usepackage{subcaption}
\usepackage{xcolor}
\usepackage{xparse}
\usepackage{hyperref}
 \usepackage{amsmath}

\usepackage[utf8]{inputenc}
\usepackage{xcolor}
\usepackage{tcolorbox}
\usepackage{lipsum}

\begin{document}

\copyrightyear{2024}
\copyrightclause{Copyright © 2024 for this paper by its authors. Use permitted under Creative Commons License Attribution 4.0 International (CC BY 4.0).}

\conference{EKAW 2024 Workshops, Tutorials, Posters and Demos, 24th International Conference on Knowledge Engineering and Knowledge Management (EKAW 2024), November 26-28, 2024, Amsterdam, The Netherlands}

\title{LLMs4Life: Large Language Models for Ontology Learning in Life Sciences}

%\tnotetext[1]{You can use this document as the template for preparing your publication. We recommend using the latest version of the ceurart style.}

\author[1]{Nadeen Fathallah}[%
orcid=0000-0001-7921-034X,
email=nadeen.fathallah@ki.uni-stuttgart.de
]
\address[1]{Analytic Computing, Institute for Artificial Intelligence, University of Stuttgart, Stuttgart, Germany}

\author[1,2]{Steffen Staab}[%
orcid=0000-0002-0780-4154,
email=steffen.staab@ki.uni-stuttgart.de]
\address[2]{University of Southampton, Southhampton, UK}

\author[3,4]{Alsayed Algergawy}[%
orcid=0000-0002-8550-4720,
email=alsayed.algergawy@uni-passau.de
]
\address[3]{Data and Knowledge Engineering, University of Passau, Passau, Germany}
\address[4]{Institute for Informatics, Friedrich-Schiller-University Jena, Jena, Germany}

%\cortext[1]{Corresponding author.}
%\fntext[1]{These authors contributed equally.}

\begin{abstract}
Ontology learning in complex domains, such as life sciences, poses significant challenges for current Large Language Models (LLMs). Existing LLMs struggle to generate ontologies with multiple hierarchical levels, rich interconnections, and comprehensive class coverage due to constraints on the number of tokens they can generate and inadequate domain adaptation. To address these issues, we extend the NeOn-GPT pipeline for ontology learning using LLMs with advanced prompt engineering techniques and ontology reuse to enhance the generated ontologies' domain-specific reasoning and structural depth. Our work evaluates the capabilities of LLMs in ontology learning in the context of highly specialized and complex domains such as life science domains.  To assess the logical consistency, completeness, and scalability of the generated ontologies, we use the AquaDiva ontology developed and used in the collaborative research center AquaDiva \footnote{\url{https://www.aquadiva.uni-jena.de/}} as a case study. Our evaluation shows the viability of LLMs for ontology learning in specialized domains, providing solutions to longstanding limitations in model performance and scalability. 

\end{abstract}


\maketitle

\section{Introduction} 
Ontology learning encompasses tasks such as ontology extraction, ontology generation, or ontology acquisition. It is the automatic or semi-automatic creation of ontologies, including extracting the corresponding domain's terms and the relationships between the concepts that these terms represent from a corpus of natural language text and encoding them with an ontology language for easy retrieval \cite{maedche2001ontology}. Ontology learning in complex domains like life sciences presents a significant challenge. Although Large Language Models (LLMs) have shown promise in automating the generation and enrichment of ontologies \cite{DBLP:conf/synasc/MateiuG23, DBLP:conf/semweb/GiglouDA23,DBLP:journals/corr/abs-2403-08345, DBLP:conf/esws/SaeedizadeB24, DBLP:journals/corr/abs-2403-05921}, their application in highly specialized domains remains difficult and understudied. The inherent complexity of specialized domains such as life sciences, coupled with domain-specific terminologies and data, limits the ability of LLMs to generate ontologies that meet the structural and logical depth required for advanced reasoning. To explore these limitations, we consider the knowledge representation and ontology developed within the collaborative research center AquaDiva. AquaDiva is a large collaborative project encompassing fields such as biology, geology, chemistry, and computer science, all working towards a shared objective of enhancing our understanding of the Earth's critical zone ~\cite{kusel2016deep}.  As the complexity and amount of data collected within AquaDiva increases, there is an increasing necessity to adopt semantic web approaches to standardize data and facilitate its integration and interoperability ~\cite{algergawy2021towards,AlgergawyTowardsSA}. To this end, the AquaDiva ontology (\textit{ADOn}) has been developed with 78.840 axioms,  8.892  concepts, and 245 object properties. We leverage the AquaDiva ontology \textit{ADOn} as a use case for evaluating the performance of our method and assessing the structural depth, logical consistency, and completeness of the generated ontologies against this established domain-specific resource.

Ontologies are critical for organizing domain knowledge in a structured, reusable way, facilitating scientific research, and supporting advanced data analysis and decision-making. In domains like AquaDiva, an accurate and logically consistent ontology can enable a better understanding of complex ecological processes, enhance data interoperability, and improve scientific communication. Current approaches to ontology learning rely heavily on manual processes, which are labor-intensive and prone to human error. Incorporating LLMs offers the potential to enhance efficiency \cite{DBLP:conf/semweb/GiglouDA23}; however, automating or semi-automating this process with LLMs requires rigorous evaluation to ensure the logical soundness, domain coverage, and adaptability of generated ontologies in complex domains \cite{DBLP:journals/corr/abs-2407-19998}. Evaluating LLMs for their ability to capture deep, complex relationships between concepts is critical, as they often produce simplified structures that lack the depth and coverage necessary for representing domain-specific intricacies. Models like Neon-GPT struggle in ontology learning for domains like life sciences due to insufficient domain adaptation; as a result, they generate ontologies with shallow hierarchies. Inadequate domain adaptation means that LLMs lack sufficient exposure to specialized training data needed to model complex domains like life sciences, leading to generic ontologies. By "shallow hierarchy," we mean that the model struggles to generate deep hierarchical structures because it has difficulty establishing "is a part of" or "is a subset of" relationships, resulting in overly simplistic ontologies with limited subclass depth. Additionally, the vast amount of information required to fully model a domain like AquaDiva often exceeds the constraints on the number of tokens LLMs can generate, leading to incomplete outputs.

In this work, we propose an evaluation-driven approach to improving ontology learning for complex domains such as life science; our approach is evaluated on the AquaDiva ontology. Our methodology evolves through an experimental evaluation pipeline, where we iteratively address the limitations identified when evaluating the generated ontologies. The contributions of this paper are as follows: 

\begin{itemize}
    \item \textbf{Extension of the NeOn-GPT Pipeline with Advanced, Domain-Driven Prompt Engineering:} We extend the NeOn-GPT pipeline by introducing advanced prompt engineering techniques driven by domain-specific requirements. This includes re-prompting strategies that iteratively refine the LLM’s output, enhancing the depth and hierarchy of the generated ontology. To further improve accuracy and adaptation, we increase the use of few-shot examples and employ advanced role-play prompting using domain-specific personas. Additionally, we develop a domain categorization strategy to handle token limitations, allowing the LLM to manage large, complex domains by breaking them into manageable subsets.
    
    \item \textbf{Introduction of Ontology Reuse in the NeOn-GPT Pipeline:} We enhance the NeOn-GPT pipeline by introducing ontology reuse, incorporating relevant existing ontological resources to evaluate reuse enhances the quality, depth, and consistency of the generated ontologies.
    
    \item \textbf{AquaDiva Ontology Case Study:} We evaluate our approach by assessing the structural complexity, depth, and logical consistency of the generated ontologies.
\end{itemize}

The paper is organized as follows: Section 2 reviews related work on ontology learning with LLMs. Section 3 details our proposed methodology. Section 4 presents the experiments that were conducted and their corresponding results, using the AquaDiva ontology as a case study. Finally, Section 5 concludes the paper and outlines future work.

\section{Related Work}

In recent years, LLMs have gained significant attention for their ability to enhance various ontology-related tasks, including ontology learning. Studies have demonstrated that LLMs can support the creation, enrichment, and refinement of ontologies, helping to automate traditionally labor-intensive tasks.  For instance, Mateiu et al. \cite{DBLP:conf/synasc/MateiuG23} leverage a fine-tuned GPT-3 to translate natural language into OWL Functional Syntax for ontology enrichment. While the approach reduces the need for manual intervention, it encounters difficulties in maintaining a deep ontology structure and avoiding the generation of irrelevant axioms. Babaei Giglou et al. \cite{DBLP:conf/semweb/GiglouDA23} propose the LLMs4OL framework, which categorizes OL tasks into three core functions: Term Typing, Taxonomy Discovery, and Non-Taxonomic Relation Extraction. Their work shows that although LLMs can perform ontology learning in a zero-shot setting, fine-tuning is necessary for domain-specific tasks, particularly for complex domains like medicine or food. Similarly, Kommineni et al. \cite{DBLP:journals/corr/abs-2403-08345} developed a semi-automated pipeline using LLMs for Competency Question (CQ) generation and Knowledge Graph (KG) construction. The system reduces human effort but highlights the need for manual validation due to the variability in LLM output and prompt sensitivity. In contrast, Saeedizade and Blomqvist \cite{DBLP:conf/esws/SaeedizadeB24} experiment with GPT-4 and other open-source models to generate OWL ontologies from ontological requirements. Their study concludes that while GPT-4 performs well in general ontology tasks, it struggles with more specialized domains, necessitating human intervention to correct errors and ensure the ontology’s completeness. Zhang et al. \cite{DBLP:journals/corr/abs-2403-05921} introduce OntoChat, a conversational framework for ontology engineering that uses LLMs to assist in requirement elicitation, CQ extraction, and ontology testing. OntoChat reduces the time and effort required for these tasks, though the authors acknowledge challenges such as LLM hallucination and the need for refinement during the ontology development process.

While these studies show the potential of LLMs in ontology learning, they share common limitations, such as shallow hierarchy generation, token limitations, and insufficient domain adaptation. For example, as highlighted by Mai et al. \cite{DBLP:journals/corr/abs-2407-19998}, off-the-shelf LLMs struggle to adapt to specialized terminologies in complex domains, and their performance is limited by token constraints and pre-existing lexical knowledge. Additionally, \cite{DBLP:journals/corr/abs-2407-19998} emphasizes the importance of integrating structured knowledge sources to improve LLM performance on domain-specific tasks. These works provide the foundation for exploring how LLMs can be effectively applied to ontology learning tasks, but further improvements are necessary to address their shortcomings. In comparison, our work improves the structural depth of generated ontologies by employing re-prompting techniques and a keyword categorization strategy to manage token constraints. Additionally, we leverage ontology reuse, incorporating existing ontological structures (such as those from ENVO) to guide the LLM in generating more detailed hierarchies and relationships. Our approach ensures consistency with established domain knowledge while allowing us to generate more comprehensive ontologies, particularly in highly specialized domains like AquaDiva ontology, where token limitations and domain specificity are critical challenges.

\section{Methodology}
\begin{figure}[!ht]
  \centering
  \includegraphics[width=17cm]{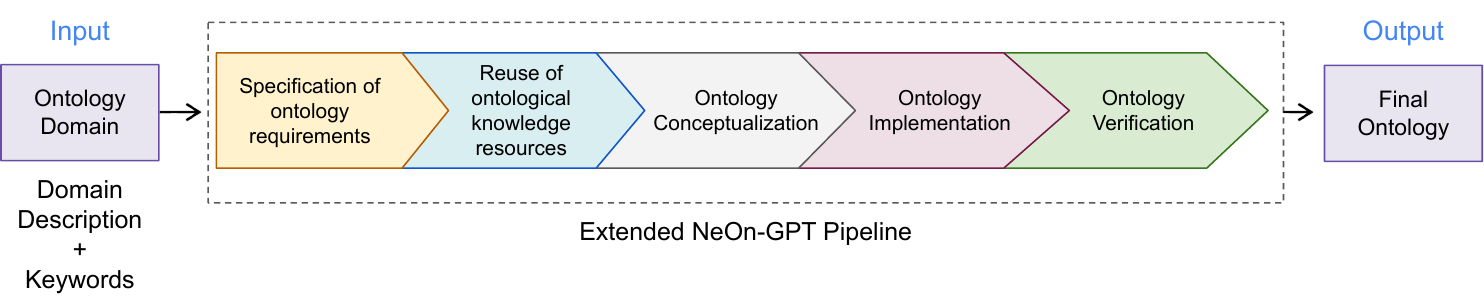}
  \caption{Overview of our proposed methodology to extend the NeOn-GPT pipeline for more complicated domains such as life science domains. The process begins with the ontology domain, incorporating domain-specific descriptions and keywords. The methodology employs a pre-trained LLM to follow a structured sequence of steps: specification of ontology requirements, reuse of ontological knowledge resources, ontology conceptualization, implementation, and verification, producing the final ontology.}
  \label{fig:method}
\end{figure}

Our approach builds on our previous work with the NeOn-GPT pipeline for ontology learning \cite{fathallah2024neon}. Using the NeOn methodology framework, we translate its structured, iterative process into a series of prompts for pre-trained LLMs, ensuring that the generated ontology is both logically sound and aligned with domain requirements. NeOn-GPT demonstrates effective ontology generation in popular domains such as wine ontology. The wine domain is widely recognized, and wine ontology serves as a benchmark in ontology learning, making it likely to have been included in the training data of pre-trained LLMs. However, our empirical experiments show that the current NeOn-GPT pipeline does not perform as well on highly specialized domains, which are often underrepresented in real-world datasets. Domains such as life sciences are scarce in the training data of language models, making them unfamiliar and challenging for the models to handle effectively. This work extends the NeOn-GPT pipeline to address the more complex and specialized domains, such as the life sciences domain, and is evaluated on the AquaDiva ontology; figure \ref{fig:method} shows the steps of the extended pipeline. Such domains require a deeper understanding of domain-specific knowledge that may not be as readily accessible to LLMs due to their complexity and relative obscurity compared to more mainstream fields. Our enhancements enable the pipeline to effectively generate ontologies in intricate domains, significantly advancing ontology learning for niche areas. 
%, and \ref{AppendixB} shows the updated prompt pipeline. 

%The first phase of the pipeline specifies the ontology requirements by using chain-of-thought prompting, guiding the model through four logical steps— defining the ontology's purpose, scope, target group, and functional requirements—that together form the complete set of ontology requirements. The original pipeline enhances the prompt by using role-play prompting, where the LLM is prompted using a persona to generate more contextually relevant responses, along with the domain description. In this work, we refine the role-play persona used in the prompts, building on our previous work that shows that more contextually enriched personas lead to more contextually relevant outputs from LLMs \cite{challenge2024}. %We used the persona shown in \ref{AppendixA}. The AquaDiva ontology description is sourced from \cite{eswc2024}. Given AquaDiva's complexity compared to the Wine domain, we anticipated limited relevant training data in the LLM. To address this, in this work, we extend the domain description with keywords and their corresponding relevant terms manually sourced from the AquaDiva ontology \footnote{\url{https://www.aquadiva.uni-jena.de/}}. These keywords, such as 'habitat' with relevant terms like 'aquifer fungi' and 'aquifer microbes,' serve as thematic anchors to guide the model’s attention to key aspects of the AquaDiva ontology.
%cite \cite{iswc2024}
\textbf{Specification of Ontology Requirements}. The first phase of the pipeline specifies the ontology requirements using chain-of-thought (CoT) prompting, guiding the model through four logical steps: defining the ontology's purpose, scope, target group, and functional requirements. To tailor the prompts for the AquaDiva domain, we integrate its description sourced from \cite{eswc2024} and manually curated keywords directly into the CoT prompt. Keywords like 'habitat' and related terms such as 'aquifer fungi' and 'aquifer microbes,' sourced from the AquaDiva ontology \footnote{\url{https://www.aquadiva.uni-jena.de/}}, act as thematic anchors, aligning the model's focus with AquaDiva-specific requirements.

In this work, we refine the role-play persona used in the prompts, building on our previous work \cite{challenge2024} that shows that more contextually enriched personas lead to more contextually relevant outputs from LLMs. 

The next step involves prompting the LLM to generate competency questions (CQ) for the ontology. To guide the model, we provide seven few-shot examples of competency questions (CQ)—an increase from previous work—including challenging properties such as "SubClassOf," which the LLM previously struggled to generate effectively. 

\textbf{Reuse of ontological knowledge resources.}  The NeOn methodology traditionally includes a step for reusing relevant ontological and non-ontological resources \cite{suarez2015neon}. However, in our earlier work, this step was omitted due to the challenges of providing large-scale ontologies to the LLM. Input token limitations made it infeasible, and chunking the content caused the model to lose context across different sections, even when chat history was retained. Additionally, when the LLM was prompted to reuse existing resources from its own knowledge base, it often hallucinated, generating non-existent references.

A key contribution of this work is introducing reuse to the NeOn-GPT pipeline based on our observations of the flat hierarchies generated by the LLM. More specifically, we identified a critical limitation in the LLM's ability to generate ontological structures that meet predefined criteria, particularly in terms of aligning with the expected count metrics derived from the gold standard AquaDiva ontology. We reused structural information (count metrics) from the AquaDiva gold standard ontology sourced from \cite{eswc2024} to prompt the LLM to improve its structural output. The prompt included specific instructions guiding the model to target predefined counts for various ontology components, such as the number of logical axioms and class, as described in \ref{two} (Experiment 2).

Although these adjustments improved the overall structure and alignment with the predefined metrics, the subclass count remained insufficient. The LLM struggled to produce a nuanced, layered hierarchy, often resulting in a flat structure with too few subclasses relative to the number of parent classes. To mitigate this, we introduced a refined prompt designed to increase the subclass count, directing the model to aim for more subclasses. We also re-prompted the model to generate a more detailed hierarchy, interconnected concepts, and robust relationships, as described in \ref{two} (Experiment 2) and \ref{four} (Experiment 4).

%. This approach fosters the creation of a more complex and interconnected hierarchical structure

To address the challenges of integrating large ontologies into the LLM, we manually extracted examples from the Environment Ontology (ENVO) \cite{ENVO}, a highly relevant resource for the AquaDiva ontology, which is the target domain in this case. This integration allowed us to assess how reusing established ontological resources improves the generated ontology’s hierarchical depth, interoperability, and relevance within the broader ontological ecosystem. In this step, the ENVO example is provided to the LLM as part of the prompt, with specific instructions on how to reuse its structure to enrich the generated ontology, as described in \ref{five} (Experiment 5). 

%throughout the generated ontology rather than limiting such richness to the example given in the prompt
%example or ref to example
\textbf{Ontology Conceptualization}
Ontology conceptualization and conceptual modeling begin with extracting entities and relationships, facilitated through few-shot prompting. Here, the LLM is guided to identify and extract entities and relationships directly from the previously generated competency questions (CQ). In this work, we modify the original prompt by incorporating domain-specific examples to tailor this process to the AquaDiva ontology. All few-shot examples are manually extracted from the AquaDiva ontology. Here's one of the few-shot examples injected in the prompt:
\begin{verbatim}
"cq1": "What measurement is associated with an observation?"
Entity: ["Observation", "Measurement"]
Property: ["hasMeasurement"]
\end{verbatim}
Following this, we prompt the model to construct a comprehensive conceptual model for the entire ontology, represented as subject-relation-object triples from the extracted entities and relations. Here's an example of the generated triples:
\begin{verbatim}
Observation -- hasMeasurement --> Measurement
\end{verbatim}

\textbf{Ontology Implementation.} In the ontology implementation phase, the original pipeline prompts the LLM to utilize the previously generated triples to create a complete ontology serialized in Turtle syntax. The original prompt emphasizes fundamental ontological constraints in the prompt, such as the correct application of Turtle syntax. The prompt also includes cautionary advice on ontology consistency and common pitfalls, such as proper prefix usage and the declaration of ontology prefixes. It provides a clear roadmap for the LLM to follow.

Next, the pipeline applies formal modeling to capture the domain's complexities and relationships within the ontology, ensuring it is logically consistent and capable of supporting advanced reasoning. This process involves prompting the LLM to introduce data properties and adjust domain and range settings accordingly, using a few-shot prompting technique. In this work, we tailor the original prompt to include domain-specific examples; here are some of the few-shot examples used:
\begin{verbatim}
:hasMeasurementValue rdf:type owl:DatatypeProperty, owl:FunctionalProperty ;
    rdfs:domain :Measurement ;
    rdfs:range xsd:float ;
    rdfs:label "has measurement value"@en .
\end{verbatim}
Subsequently, the formal modeling process includes object properties, such as inverse, reflexive, transitive, symmetric, and functional properties. The prompts are structured to ensure that these object properties are meaningfully integrated across the entire ontology rather than being limited to isolated snippets. Building on insights from our empirical experiments, in this work, we implemented syntax consistency restrictions and cautionary advice on ontology consistency and common pitfalls across all prompts, not just the initial one. This decision arose from observations that inconsistencies often appeared in later stages of ontology generation when restrictions were only applied initially. Additionally, we observed the tendency of the LLM to adopt Occam's razor approach when correcting inconsistencies during the \textbf{Ontology Verification} phase, often leading to the omission of crucial properties or classes. To counter this, we guided the LLM to generate a syntactically correct and logically consistent ontology from the outset, reducing the need for later corrections.

To enhance the ontology's usability and readability, the original prompt pipeline includes instructions for enriching entities and relationships with natural language descriptions and adding essential metadata such as IRIs, labels, and versioning information. 

Additionally, a few-shot prompt populates the ontology with real-world instances, grounding the ontology in practical data and facilitating knowledge discovery. We adapt these examples to align with our specific domain. Here is a sample of the few-shot examples used:
\begin{verbatim}
:Exbio_Antibodies rdf:type :Company, owl:NamedIndividual ;
    rdfs:label "Exbio Antibodies"@en .

:Becton_Dickinson_BD_Biosciences rdf:type :Company, owl:NamedIndividual ;
    rdfs:label "Becton Dickinson (BD Biosciences)"@en .
\end{verbatim}

To further improve the structural hierarchy, depth of the ontology, and ontology coverage, we employed re-prompting. Corrective re-prompting is a 
technique in prompt engineering where an LLM is asked the same question again to improve the quality of its responses using error-related feedback \cite{raman2022planning, challenge2024}. We use this approach to improve the initial output from the LLM. This iterative process involved prompting the LLM again to refine and extend the ontology, specifically emphasizing increasing the subclass count and creating a more layered hierarchy, as described in \ref{four} (Experiment 4).

\textbf{Ontology Verification} Following ontology generation, the NeOn-GPT workflow employs RDFLib for syntax validation, HermiT and Pallet reasoners for consistency checking, and a custom-built pitfall detection module for identifying common ontology issues. Errors and inconsistencies identified by these tools are used to prompt the LLM for corrections, ensuring the ontology is both syntactically sound and logically coherent. 

\section{Experiments and Results}
In this section, we present a series of experiments and their corresponding results to evaluate the LLM's performance before and after updating the NeOn-GPT pipeline; the generated ontologies are evaluated in terms of logical consistency and structural depth. Our objective is to assess how the proposed updates impact the LLM’s ability to generate ontologies for complex life science domains, specifically using AquaDiva ontologies. The AquaDiva ontology domain encompasses the study of groundwater ecosystems, integrating hydrogeology, microbial ecology, geochemistry, karst systems, and environmental science. This ontology supports the annotation and standardization of diverse datasets related to subsurface habitats. The current AquaDiva ontology has 78.840 axioms, 8.892 concepts, and 245 object properties \cite{eswc2024}. All experiments are conducted using GPT-4o \cite{openai2024gpt4o}. The results of these experiments are discussed to illustrate the improvements achieved through the updated pipeline. Our code base is publicly available for research and development
purposes, accessible at: \url{https://github.com/NadeenAhmad/NeOn-GPTAquaDivaOntology}. It includes all details about the prompts, interactions with LLMs, and the methodology used in our approach.

\subsection{Experiment 1: Baseline NeOn-GPT (AquaDiva)} \label{one}

In this experiment, we applied the original NeOn-GPT pipeline without any of the enhancements introduced in this work. The only modification was the inclusion of 222 domain-specific keywords alongside the textual descriptions in the input to the LLM to compensate for the anticipated scarcity of relevant training data related to AquaDiva. This allowed us to evaluate the LLM's performance in its original configuration when applied to a complex life sciences domain.
\subsubsection{Results of Experiment 1: Baseline NeOn-GPT (AquaDiva)}
\begin{figure}[!ht]

  \centering
  \begin{minipage}{0.6\textwidth}
    \centering
    \includegraphics[width=\linewidth]{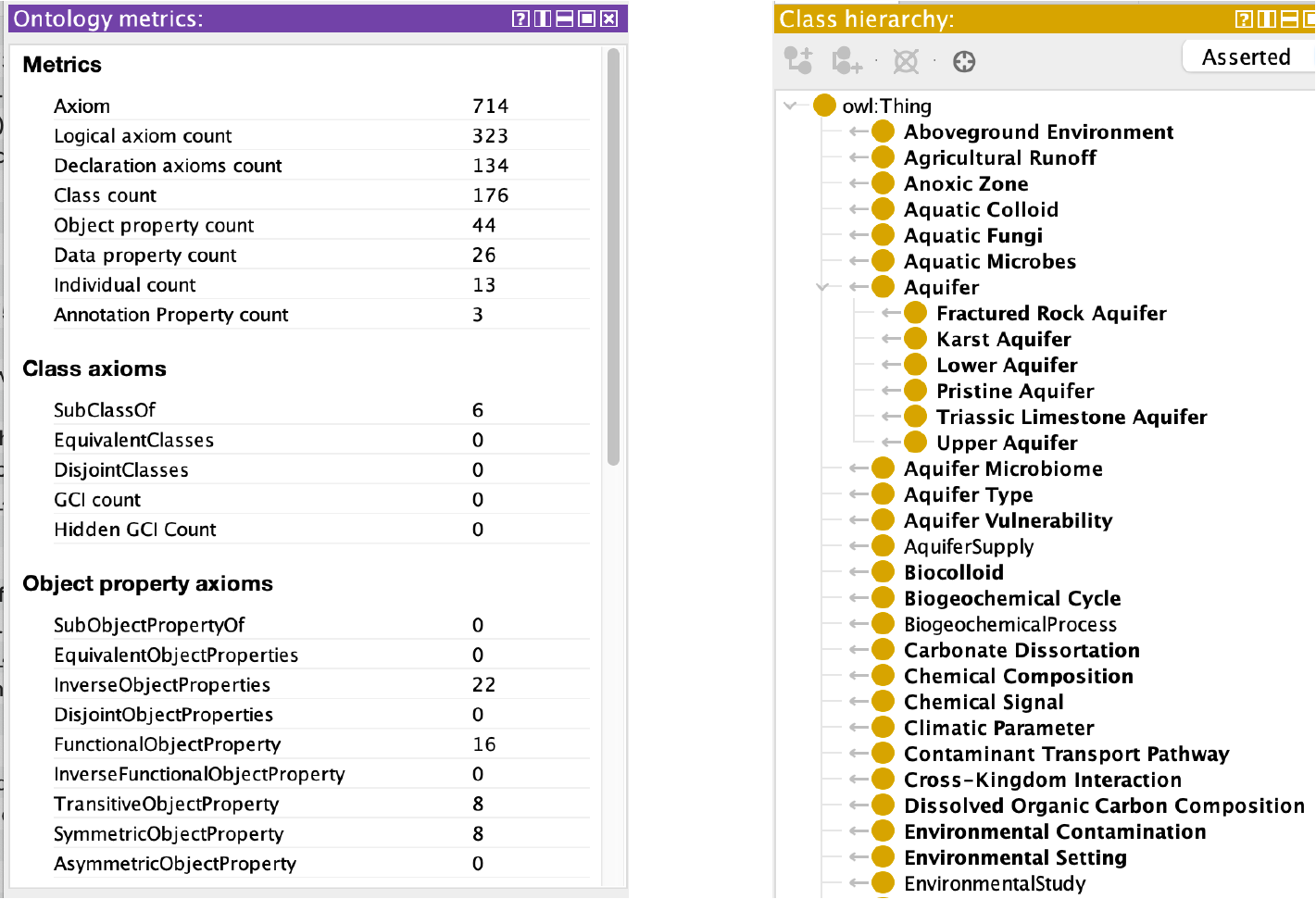}
   
  \end{minipage}%
  \hfill
  \begin{minipage}{0.4\textwidth}
    \centering
    \includegraphics[width=\linewidth]{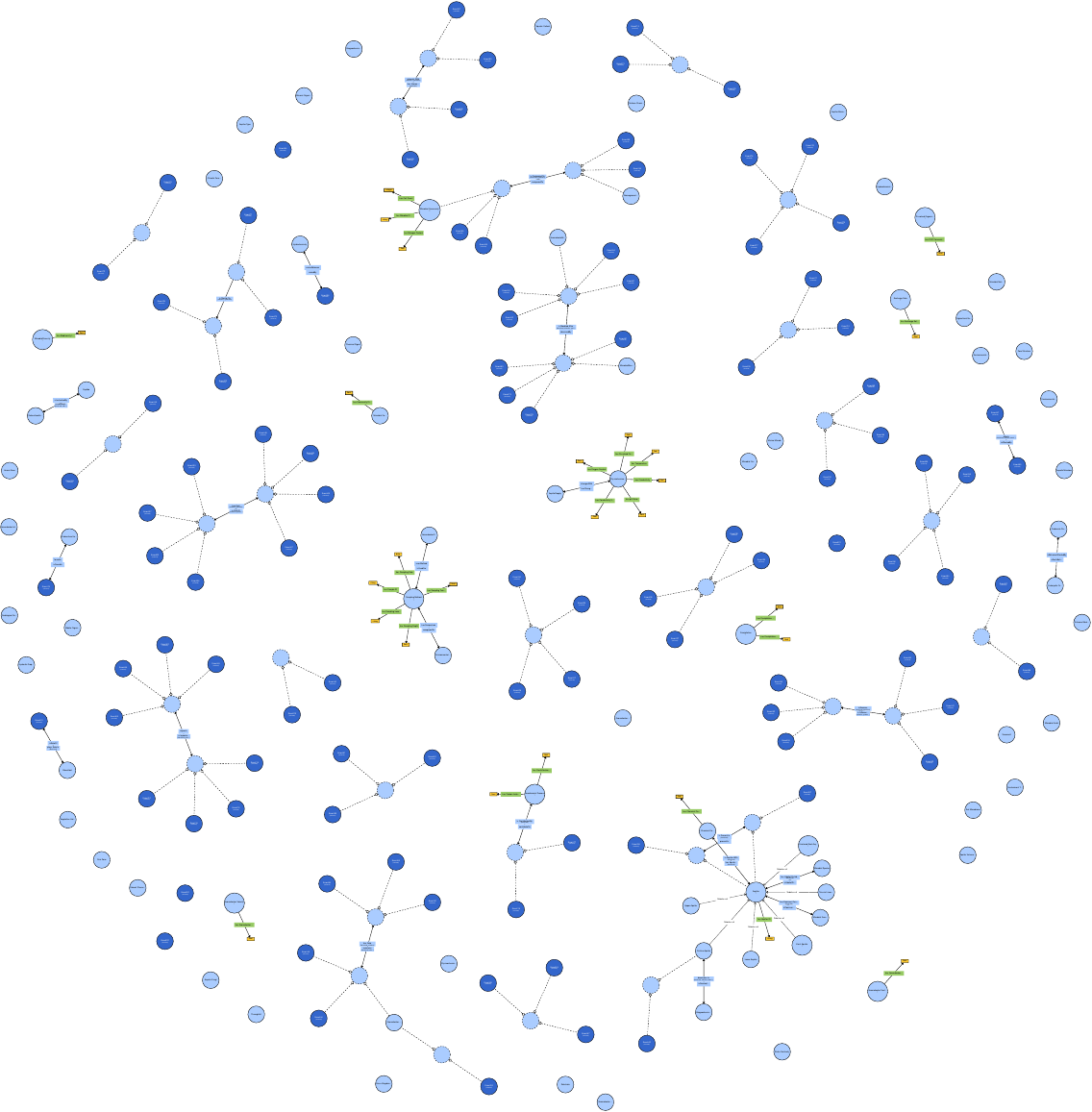}
    
  \end{minipage}
  \caption{Visualization of the AquaDiva ontology generated from Experiment 1. The left side presents ontology metrics. The center panel shows a portion of the hierarchy of classes and their relationships (visualized using Protégé), while the right side features a structural network representation of the ontology generated using WebVOWL 1.1.7.}
   \label{fig:Experiment1Res}
\end{figure}

In this experiment, we evaluated the LLM-generated AquaDiva ontology against the AquaDiva gold standard ontology \cite{eswc2024}. While the LLM successfully captured key concepts such as 'aquifers' and 'microbial communities', the ontology remained overly simplistic, with sparse hierarchy, and lacked the complexity needed for advanced ecological modeling. The metrics and class hierarchy from the newly generated ontology are shown in Figure \ref{fig:Experiment1Res}. Compared to the gold standard, the LLM-generated ontology included 176 classes (significantly fewer than the 8,892 in the gold standard) and only 44 object properties, which resulted in the omission of crucial relationships and subclass hierarchies. The absence of equivalent and disjoint classes, as well as a reduced set of logical axioms (323 versus 16,303 in the gold standard), further limited its ability to accurately represent the interactions within ecological systems.

Despite including important concepts like "Aquifer" and its subclasses (e.g., "Fractured Rock Aquifer," "Karst Aquifer") and environmental factors such as "Aquifer Vulnerability" and "Biogeochemical Cycle," the generated ontology lacked the relational depth necessary to describe the interactions between these entities. This omission impacts the ability to model and reason about the relationships, such as the specific types of microbial interactions within these environments. Additionally, the reduced number of object properties, individuals, and data properties (13 individuals and 26 data properties in total) further simplified the representation, preventing complex ecological relationships and taxonomic structures from being expressed. The simplified logical framework, including only 6 SubClassOf axioms, made it difficult to support detailed ecological queries, significantly reducing its utility for in-depth reasoning about environmental and biological phenomena.

\subsection{Experiment 2: Count Metric-Guided Prompts (AquaDiva)} \label{two}
In Experiment 2, we addressed the limitations identified in \ref{one} (Experiment 1) by revising the prompt pipeline to incorporate explicit count metrics from the AquaDiva gold standard, such as the number of classes (8,892) and object properties (245). The prompt instructed the LLM to align with these metrics, emphasizing a subclass count of at least n-1 (where n is the total number of classes) to address shallow hierarchies observed in \ref{one} (Experiment 1).

\subsubsection{Results of Experiment 2:  Count Metric-Guided Prompts (AquaDiva)} 
\begin{figure}[!ht]

  \centering
  \begin{minipage}{0.6\textwidth}
    \centering
    \includegraphics[width=\linewidth]{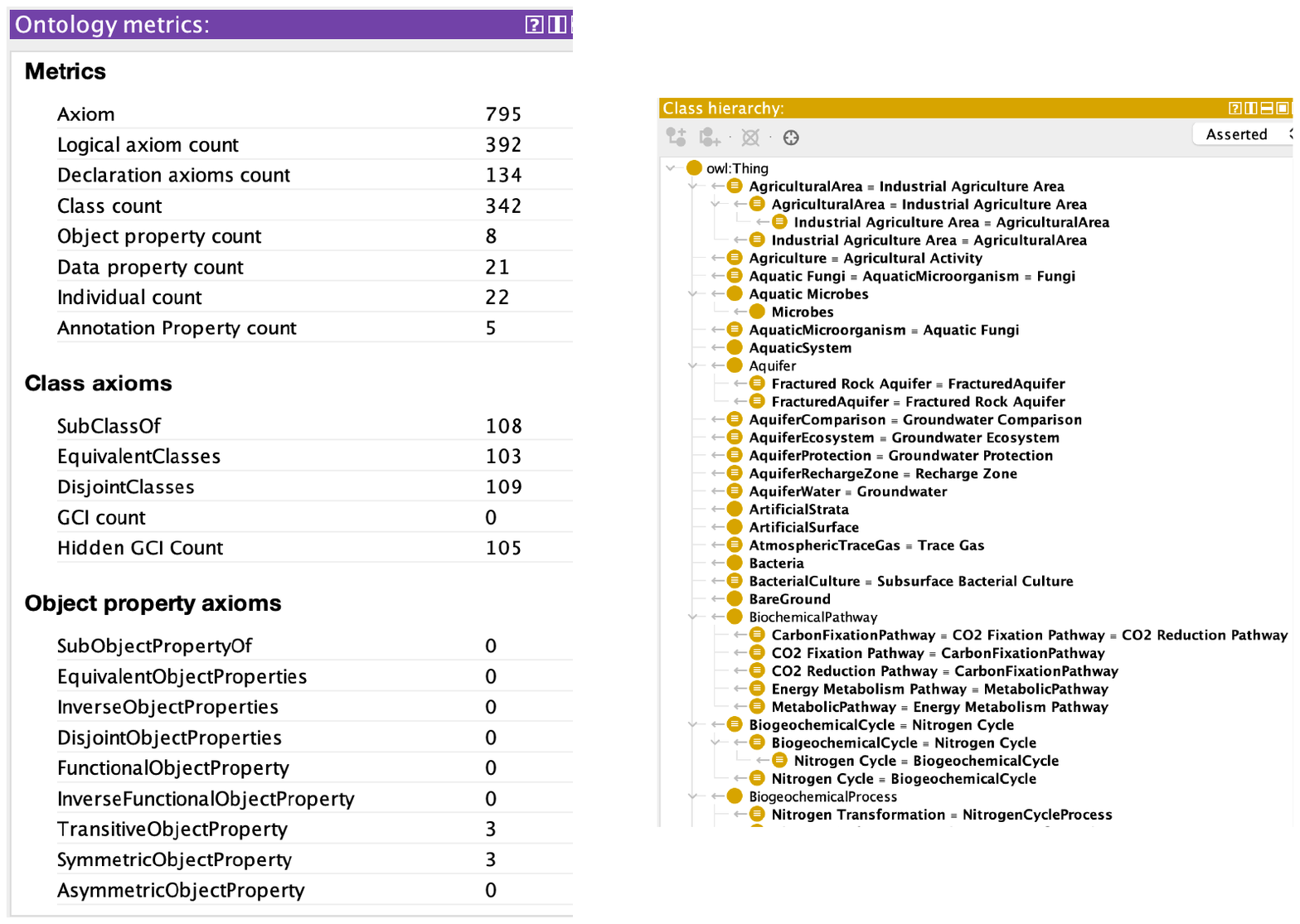}
   
  \end{minipage}%
  \hfill
  \begin{minipage}{0.4\textwidth}
    \centering
    \includegraphics[width=\linewidth]{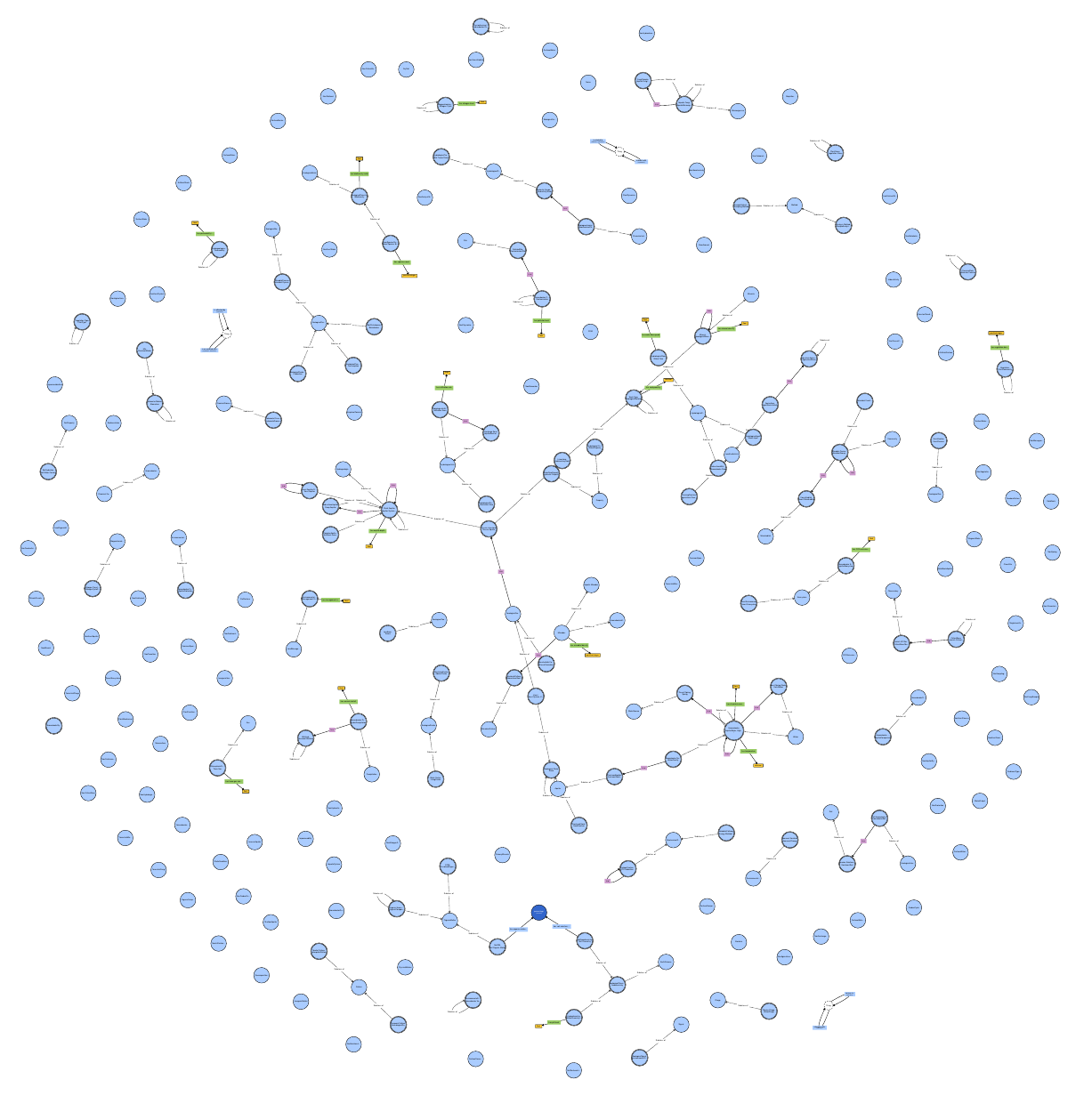}
    
  \end{minipage}
  \caption{Visualization of the AquaDiva ontology generated from Experiment 2.}
   \label{fig:Experiment2Res}
\end{figure}

The ontology generated in Experiment 2 demonstrated significant improvements over the initial version, exhibiting a more interconnected structure with increased density and a more layered hierarchy. As shown in Figure \ref{fig:Experiment2Res}, this version includes 342 classes and 795 axioms, a notable increase from Experiment 1 but still lower than the expected 8,892 classes and 78,840 axioms in the AquaDiva gold standard ontology. This iteration represents a broader range of concepts and relationships, demonstrating more alignment with the domain's complexity. For instance, the ontology now contains 108 SubClassOf axioms and 103 EquivalentClasses axioms, improving upon \ref{one} (Experiment 1), resulting in a more layered hierarchy with more subclass levels like (e.g., “HydroChemistry” – “SubClassOf” → “Geological Chemistry” – “SubClassOf” → “Earth Science”). However, the ontology still falls short in certain areas, particularly in object property count, which remains at 8 compared to the expected 245 in the gold standard. This can be partially attributed to GPT-4o's limitations, including its 4096-token output limit \cite{openai2024gpt4o}, which restricts the amount of content generated in a single response. Additionally, the LLM's mathematical limitations, particularly in precise counting tasks \cite{DBLP:conf/nips/FriederPCGSLPB23}, likely contribute to the discrepancies in class and axiom counts. Moreover, some redundancy persists, with overlapping object properties such as "interact with" and "interacts with," indicating a need for further refinement.

\subsection{Experiment 3: Merging Ontologies (AquaDiva)} \label{three}
In Experiment 3, we merged the ontologies generated from Experiments 1 and 2. Merging ontologies is a widely accepted approach in ontology engineering, as it can improve coverage, coherence, and the overall quality of the resulting ontology by combining complementary strengths from different sources \cite{de2006ontology}. By merging ontologies, we can address gaps that may exist in individual outputs and ensure a more comprehensive and robust knowledge representation. We utilized the RDFLib library to merge these ontologies; the ontology from \ref{two} (Experiment 2), which contains richer concepts and better hierarchical structures, was used as the foundation. Unique and complementary elements from \ref{one} (Experiment 1), particularly object properties and relationships, were incorporated to enhance depth and coverage.
\subsubsection{Results of Experiment 3: Merging Ontologies (AquaDiva)}
\begin{figure}[!ht]

  \centering
  \begin{minipage}{0.6\textwidth}
    \centering
    \includegraphics[width=\linewidth]{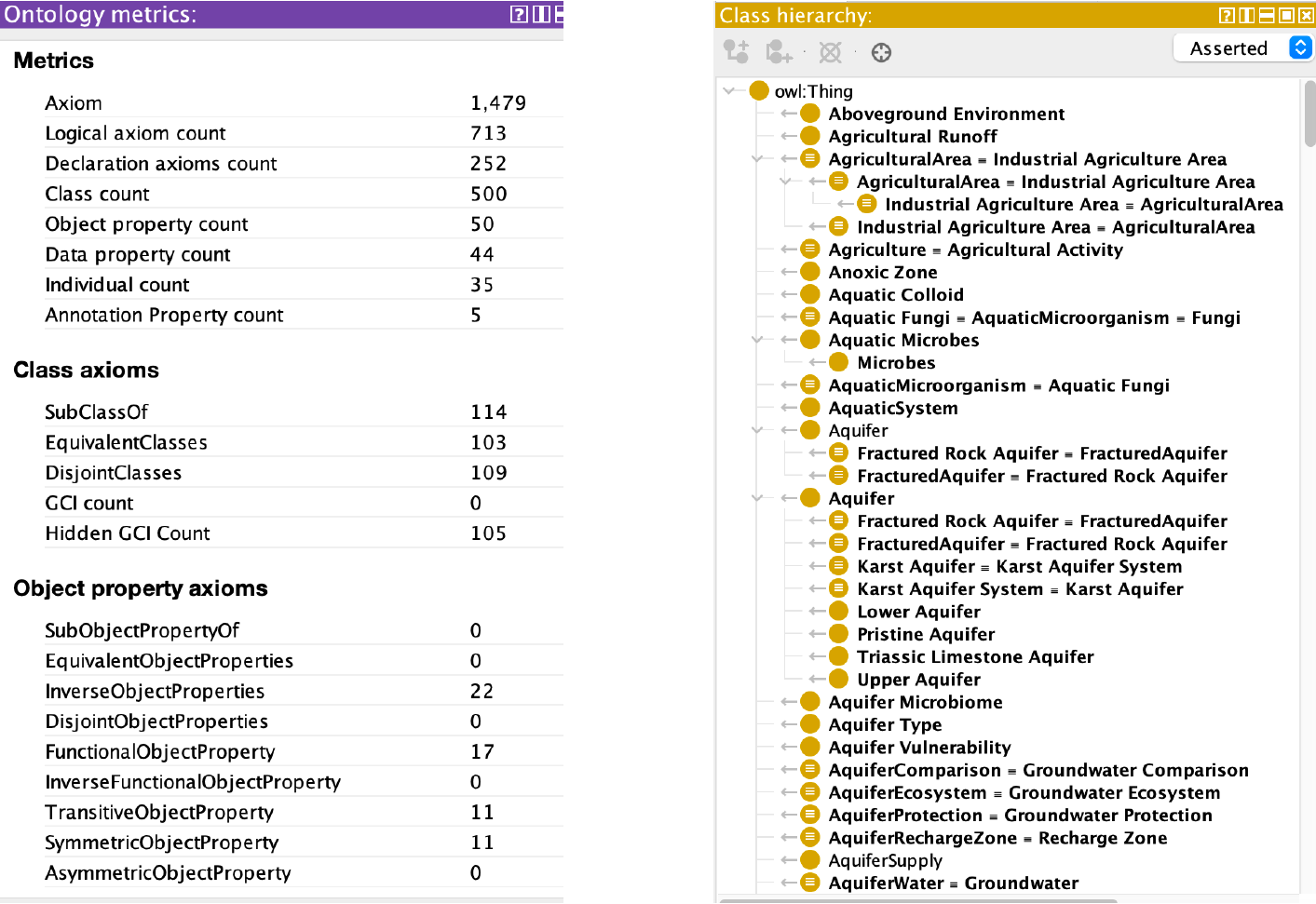}
   
  \end{minipage}%
  \hfill
  \begin{minipage}{0.4\textwidth}
    \centering
    \includegraphics[width=\linewidth]{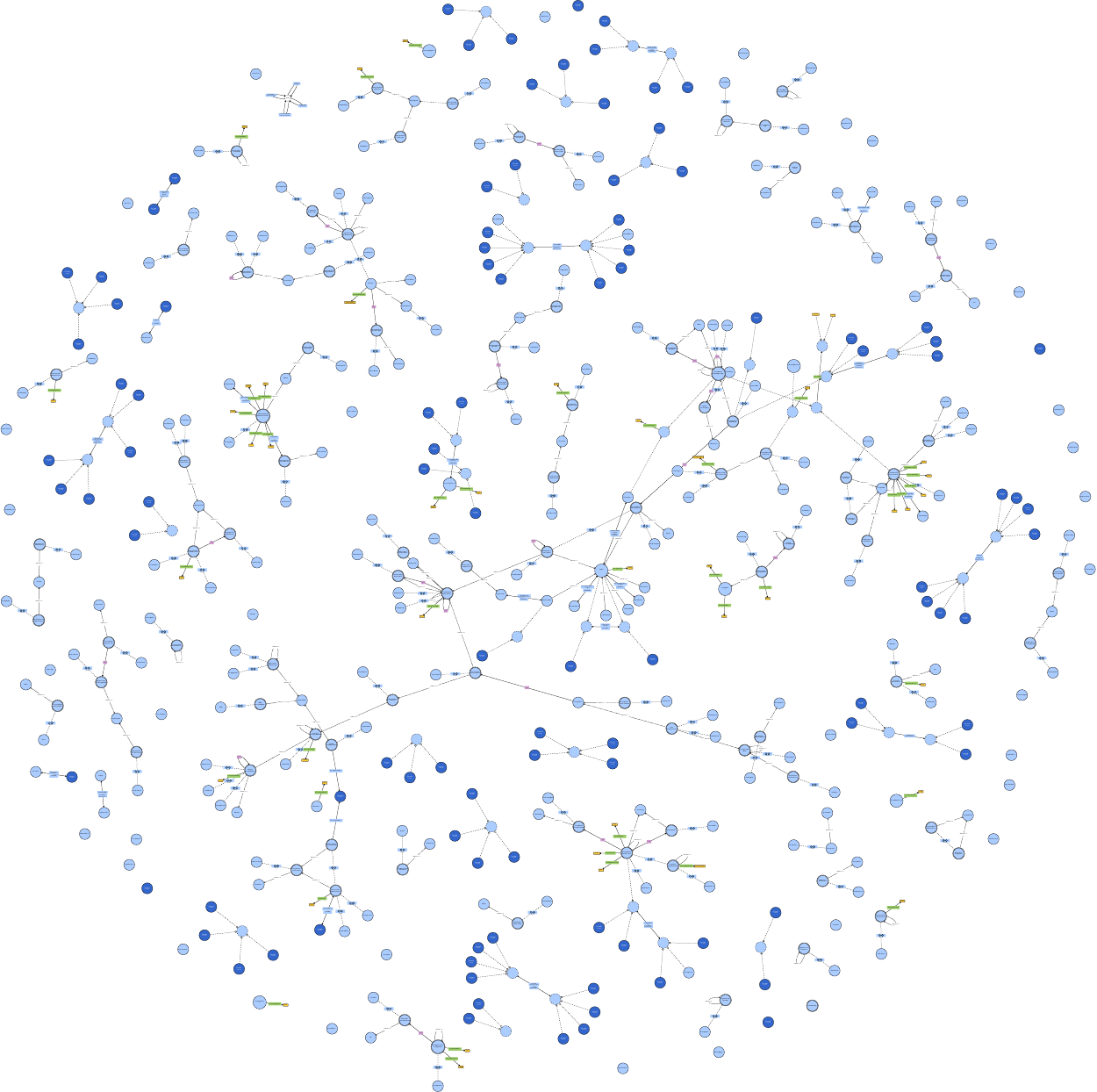}
    
  \end{minipage}
  \caption{Visualization of the AquaDiva ontology generated from Experiment 3.}
   \label{fig:Experiment3Res}
\end{figure}

The ontology generated in Experiment 3 shows a notable improvement in several key metrics, as shown in \ref{fig:Experiment3Res}. The total axiom count increased to 1,479, and the object property count rose to 50, compared to the lower counts observed in earlier experiments. These increases suggest that merging ontologies effectively captured a broader set of relationships and axioms, resulting in a more comprehensive ontology.
However, while the merged ontology shows progress, some limitations persist. The class count, now at 500, is significantly improved compared to previous versions but remains below the gold standard AquaDiva ontology. There are still discrepancies between the generated metrics and the expected counts, particularly regarding data and annotation properties, where further refinement is needed to match the complexity of the domain. While the object property count has risen, it still falls short of the expected 245, indicating that additional adjustments are necessary to fully capture the domain's complexity.
Notably, the ontology has made significant strides in logical consistency, with 713 logical axioms, and includes 114 SubClassOf axioms. This improved structure provides better support for defining relationships such as hierarchical taxonomies and equivalence between classes (e.g., "Aquatic Fungi" = "Aquatic Microorganism = Fungi"). Yet, despite these advances, the number of disjoint classes (109) still lags, impacting the ontology's ability to distinctly differentiate between overlapping or mutually exclusive categories, which is critical for accurate environmental modeling.

\subsection{Experiment 4: Re-prompting \& Advanced Role-play Prompting (Habitat)} \label{four}
%Enhanced Pipeline Using Re-prompting to Generate an Ontology for the Habitat Category within the AquaDiva ontology
In previous experiments, we tried to avoid exceeding the output token limitation of LLMs by instructing the model in each prompt to print only the new parts of the ontology. We manually aggregated these responses into a single ontology to prevent the model from regenerating the entire ontology repeatedly. However, this approach alone was not enough to fully overcome the token limitation issue. In Experiment 4, we addressed the limitations of using GPT-4o for generating a comprehensive ontology due to the model’s output token constraints. Rather than attempting to generate the entire AquaDiva ontology at once, we instructed the LLM to categorize 222 AquaDiva-specific keywords into distinct groups, resulting in 22 categories. Inspired by the improvements observed in \ref{three} (Experiment 3), where merging outputs led to a more interconnected ontology, we envision that this categorization will help generate better ontologies for each category. Ultimately, merging these individual ontologies will result in a larger and more comprehensive representation of the AquaDiva ontology.

We selected the "Habitat" category due to its ecological significance in AquaDiva. By developing an ontology for this category, we aimed to create a detailed and accurate representation that could serve as a model for expanding to other categories. To improve the quality and precision of the generated ontology, we applied several enhancements to the prompt pipeline in \ref{two} (Experiment 2). First, we provided the LLM with a detailed description of the Habitat category along with relevant keywords, ensuring a richer domain-specific context as input. Additionally, we increased the number of few-shot examples from three examples to seven examples, tailoring them to the specific concepts within the Habitat domain. Furthermore, we refined the role-play persona used in the prompts, building on our findings that show enriched personas can yield higher-quality outputs \cite{challenge2024}. We refined the role-play persona to represent an expert aquatic ecologist, leveraging domain knowledge to guide the model more effectively. The persona provided detailed instructions on how to structure the Habitat ontology with rich ecological context, ensuring domain relevance.

To iteratively refine the generated ontology, we applied re-prompting, asking the model to enhance the hierarchical depth and align with predefined metrics after the initial output. For example, a prompt to increase the subclass count to at least n-1, where n refers to the total number of classes, while other prompts address shallow hierarchies.
%To iteratively refine the generated ontology, we applied re-prompting, asking the model to enhance the hierarchical depth and align with predefined metrics after the initial output. For example, prompts focused on increasing the subclass count (n-1 relationship) and other prompts to address shallow hierarchies.

%After the initial generation, we re-prompted the LLM to refine the ontology, allowing for iterative improvements. This approach follows best practices in prompt engineering, where multiple iterations help fine-tune and improve the model’s output to better match the expected metric counts \cite{raman2022planning, challenge2024}.

%To further guide the LLM, we reused explicit target metrics from the AquaDiva gold standard ontology, as done in \ref{two} (Experiment 2), to better align the generated ontology with predefined metrics.
\subsubsection{Results of Experiment 4: Re-prompting \& Advanced Role-play Prompting (Habitat)} 
\begin{figure}[!ht]

  \centering
  \begin{minipage}{0.6\textwidth}
    \centering
    \includegraphics[width=\linewidth]{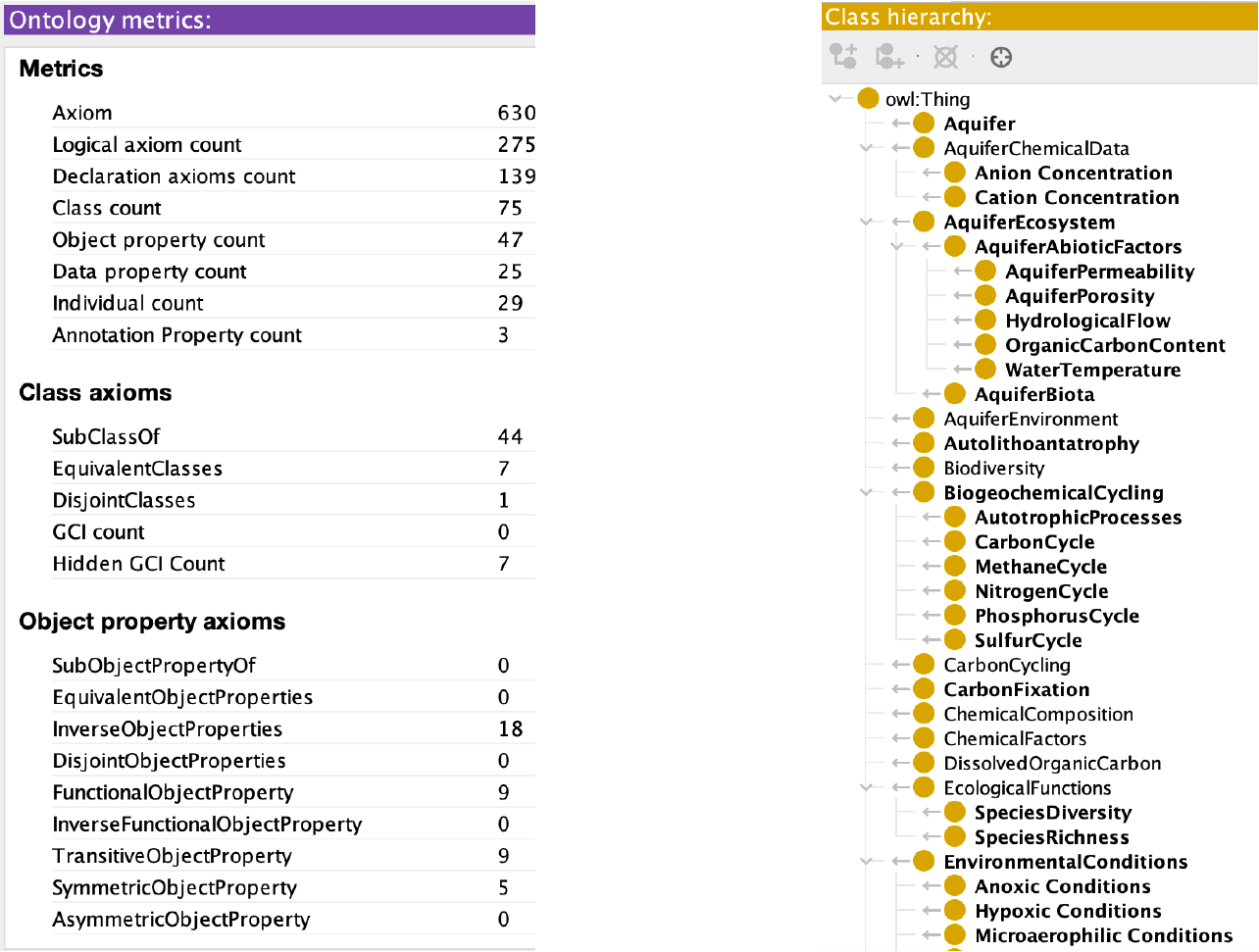}
   
  \end{minipage}%
  \hfill
  \begin{minipage}{0.4\textwidth}
    \centering
    \includegraphics[width=\linewidth]{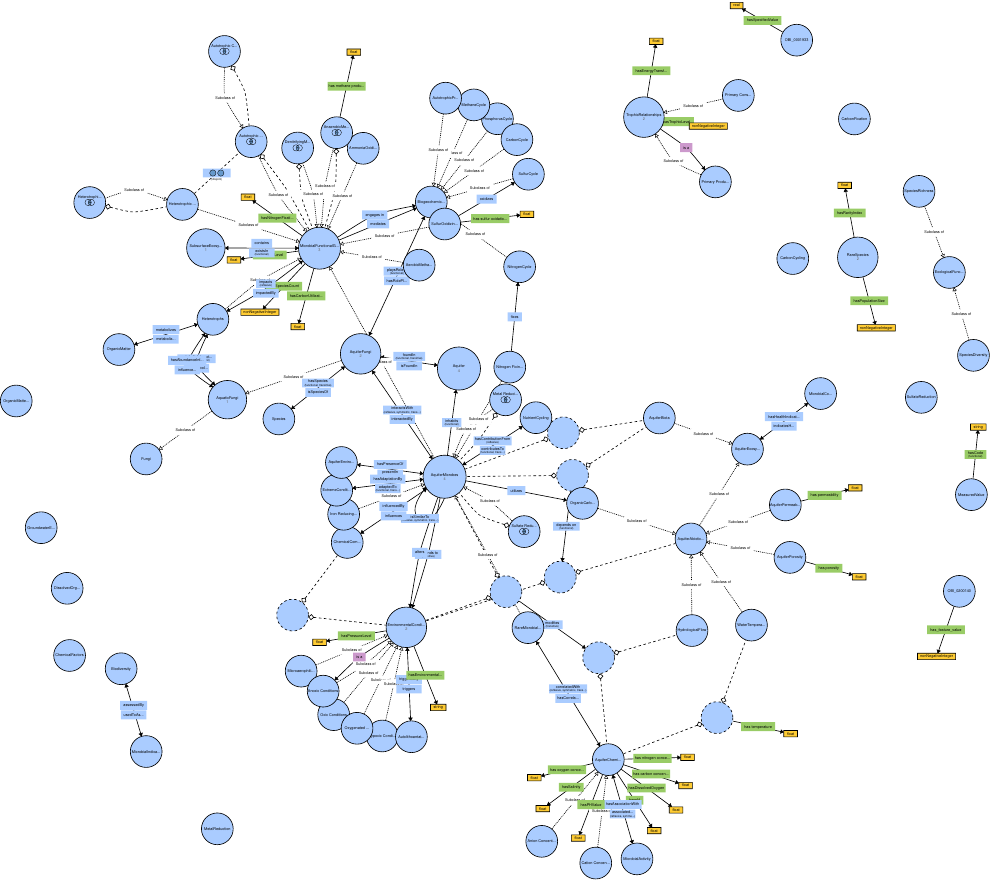}
    
  \end{minipage}
  \caption{Visualization of the Habitat ontology within the AquaDiva ontology generated from Experiment 4.}
   \label{fig:Experiment4Res}
\end{figure}

The results from Experiment 4 show that while directing the LLM's attention to the 'Habitat' category allowed for a more concentrated development of the ontology, certain limitations remain evident. The ontology is shown in Figure \ref{fig:Experiment4Res} shows part of the ontology metrics and class hierarchy. The ontology generated has a total of 630 axioms, 275 logical axioms, and 75 classes. Although these metrics indicate progress, particularly in the object property count (47), they still fall short in several areas. For instance, there is only a single DisjointClasses axiom and 3 EquivalentClasses axioms, reflecting an incomplete structure in terms of class relationships. Moreover, the number of SubClassOf axioms (44) remains insufficient for a fully detailed hierarchical structure, and the ontology still lacks comprehensive disjointness and equivalence axioms, which are crucial for distinguishing and relating different categories within the ecological domain.

\subsection{Experiment 5: Reuse (Role)} \label{five}

In Experiment 5, we generated an ontology for the "Role" category within the AquaDiva ontology with the same enhancements done to the pipeline in \ref{four} (Experiment 4). To address the lack of hierarchical depth observed in previous experiments, we improved the subclass structure by incorporating an ontology reuse strategy using a detailed example manually extracted from the ENVO ontology. This reuse example demonstrated an extensive hierarchy of classes and subclasses, utilizing a visual structure with arrows to represent increasing levels of subclass specificity. This model served as a guide for the LLM, ensuring that each class in the Role ontology would have a well-defined hierarchy of subclasses, thereby enhancing the overall depth and complexity of the ontology. Here's a simplified portion of the example provided in the prompt for reuse to illustrate the hierarchical structure:
\begin{verbatim}
-> biological_process
--> biodegradation
--> cellular process
---> cellular metabolic process
----> cellular alkane metabolic process
----> photosynthesis
\end{verbatim}

In this format, each arrow represents increasing levels of subclass hierarchy, starting from broad categories like "biological\_process" and moving down to more specific entities such as "cellular\_process." This reuse example, manually curated from ENVO, helped guide the LLM in generating deeper subclass hierarchies and producing a more layered structure in the Role ontology.

\subsubsection{Results of Experiment 5: Reuse (Role)} 
\begin{figure}[!ht]

  \centering
  \begin{minipage}{0.6\textwidth}
    \centering
    \includegraphics[width=\linewidth]{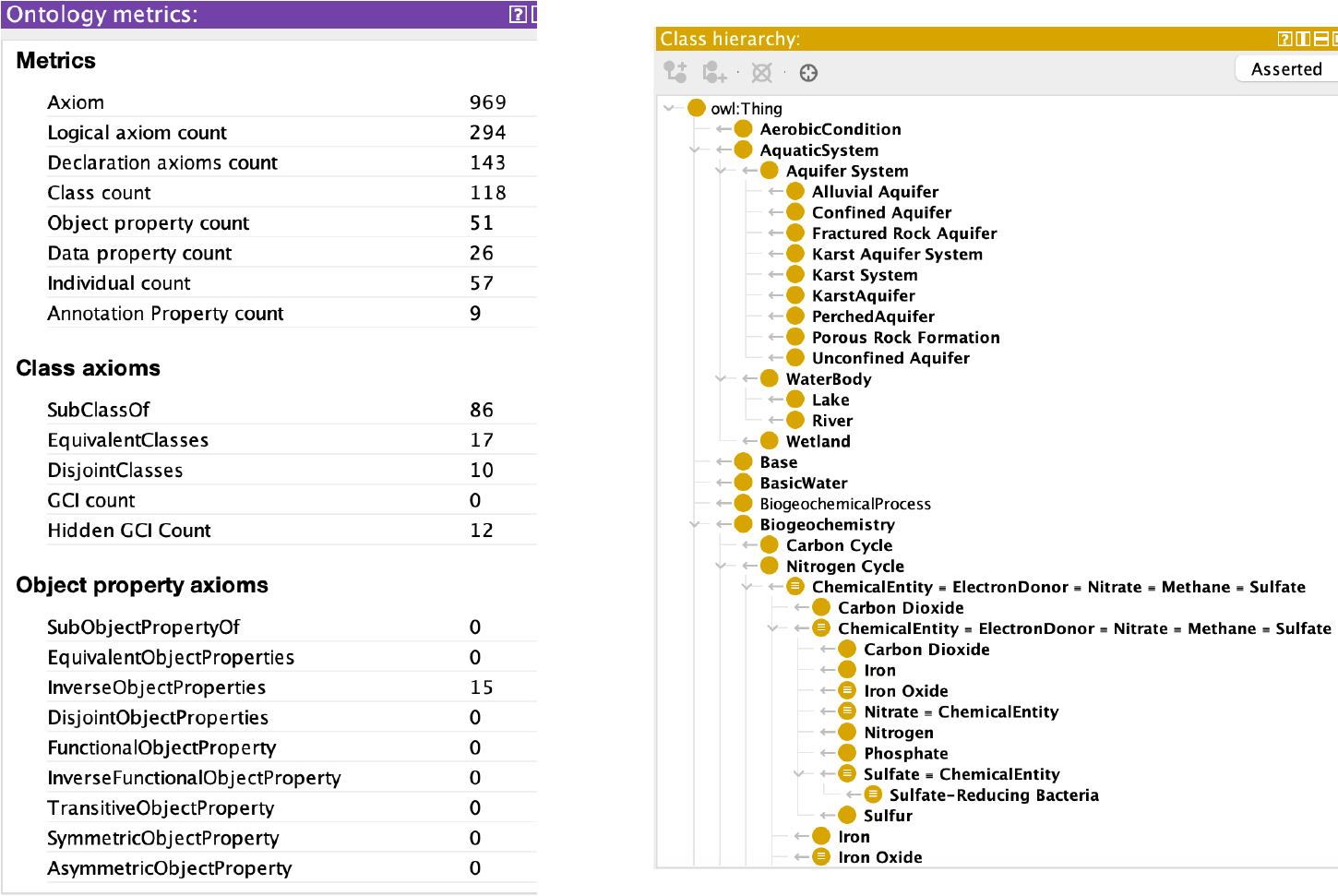}
   
  \end{minipage}%
  \hfill
  \begin{minipage}{0.4\textwidth}
    \centering
    \includegraphics[width=\linewidth]{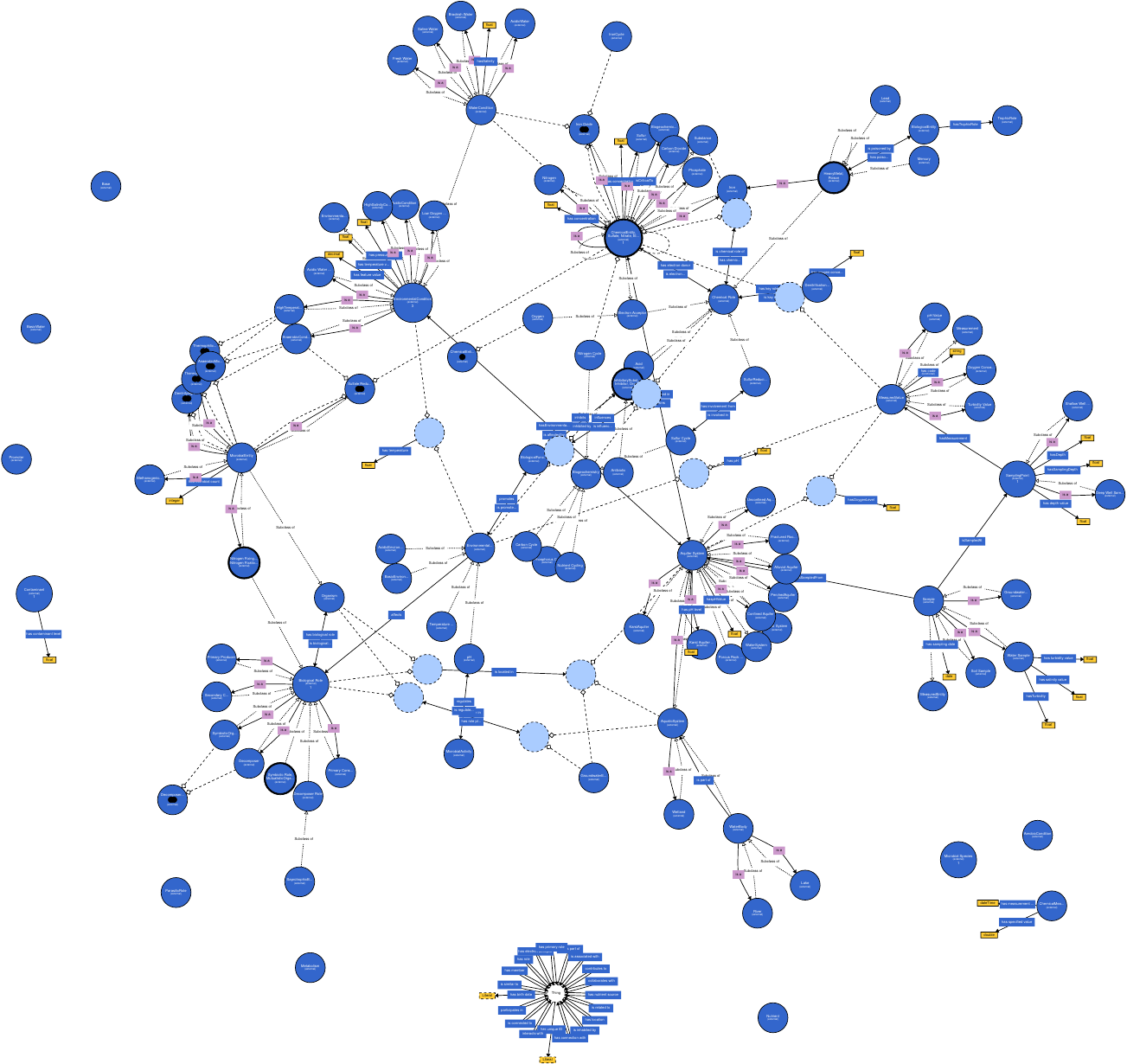}
    
  \end{minipage}
  \caption{Visualization of the Role ontology within the AquaDiva ontology generated from Experiment 5.}
   \label{fig:Experiment5Res}
\end{figure}

The role ontology generated in Experiment 5 demonstrates notable strengths, particularly in its axiom count, which includes 969 axioms, and class count, which includes 118 classes. These metrics accurately represent the relationships within the 'Role' domain, showcasing the ontology's potential for supporting complex reasoning tasks. Additionally, the inclusion of 57 individual instances suggests a more comprehensive and practically applicable ontology, contributing to its overall depth and usability for modeling in the AquaDiva ontology. The ontology is shown in Figure \ref{fig:Experiment5Res}, which highlights part of the ontology metrics and class hierarchy.

One of the key improvements in this experiment was the significant increase in the subclass count compared to \ref{four} (Experiment 4), with the ontology now containing 86 subclasses. This improvement was achieved through the incorporation of the manually extracted ENVO example, which provided a structured reuse of existing hierarchical ontologies. The enhanced subclass hierarchy contributed to a more layered and detailed ontology, addressing prior limitations in the structural depth observed in earlier experiments.

However, despite these strengths, the ontology still exhibits significant limitations; while there is some improvement in logical consistency (e.g., 17 EquivalentClasses), the ontology remains underdeveloped in terms of disjoint class distinctions, with only 10 DisjointClasses axioms. This gap weakens its logical coherence and reduces its robustness for reasoning tasks. Moreover, the broad and generic nature of some classes within the 'Role' category, such as "Biological Role" or "Chemical Role," raises concerns about the potential inclusion of overly generic terms that could dilute the ontology's focus and reduce its utility in the specific context of the AquaDiva ontology.

\subsection{Experiment 6: Reuse of domain-specific examples
(Carbon \& Nitrogen Cycling)}
%Enhanced Pipeline Using Domain-specific Ontology Reuse to Generate an Ontology for the Carbon and Nitrogen Cycling Category within the AquaDiva ontology
In Experiment 6, we generated an ontology for the Carbon and Nitrogen Cycling domain, building on the lessons learned from previous experiments. Earlier attempts demonstrated that the reuse of existing ontological resources can significantly improve terminology generation and result in a more complex and layered hierarchy. This was evident in the increase in the number of classes and subclasses from \ref{four} (Experiment 4) to \ref{five} (Experiment 5). Motivated by these findings, we selected the Carbon and Nitrogen Cycling domain to evaluate how the reuse of an example with domain-specific terminology could further enhance the ontology generation process. 

This experiment incorporates all the improvements added to the original NeOn-GPT prompt pipeline. First, we continued using the advanced role-play persona from \ref{four} (Experiment 4) and \ref{five} (Experiment 5) to maintain contextual relevance. For this domain, we provided a detailed description along with domain-specific keywords to guide the model's understanding. Additionally, we increased the number of few-shot examples, tailoring them to the Carbon and Nitrogen Cycling domain. To ensure logical consistency and structural depth, we implemented syntax and consistency restrictions at all stages of ontology generation prompts, reducing the need for later corrections related to missing properties or classes.

We enhanced the reuse of existing ontological resources. Instead of using broader, generic examples, we incorporated specific components manually extracted from ENVO that closely align with the Carbon and Nitrogen domain. This targeted reuse approach provided a clearer structure, ensuring the ontology reflected accurate hierarchical depth, interconnected concepts, and detailed relationships. A portion of the reuse example included classes like "carbon atom" and "nitrogen atom" and their corresponding subclasses, organized into multiple levels of hierarchy.

Here's a simplified portion of the example provided in the prompt for reuse to illustrate the hierarchical structure:
\begin{verbatim}
   -> carbon_atom
   --> carbon-13_atom
   --> carbon-14_atom
   -> dissolved_carbon_atom_in_environmental_material
   --> dissolved_carbon_atom_in_soil
   --> dissolved_carbon_atom_in_water

\end{verbatim}
This example, manually curated from ENVO, demonstrated the expected level of hierarchy, with each class and its corresponding subclasses represented by increasing levels of specificity.

\subsubsection{Results of Experiment 6: Reuse of domain-specific examples
(Carbon \& Nitrogen Cycling)} The Carbon and Nitrogen Cycling ontology developed in Experiment 6 shows significant improvements in capturing complex biochemical processes. Key entities such as "Carbon Fixation," "Nitrogen Transformation," and "Methanogenesis" are accurately modeled, with 157 classes and 63 object properties, including 13 functional, 10 symmetric, and 10 transitive properties, enabling detailed representations of interactions like fixing nitrogen and releasing methane.

A major improvement is the hierarchical depth, with 130 SubClassOf axioms, enhanced by reusing domain-specific components from the ENVO ontology. This includes entities such as "ammonia oxidation"  and "biogeochemical cycle", reflecting a richer, more structured subclass hierarchy. The ontology also includes 1,169 axioms, 455 of which are logical, offering a more detailed representation of processes like "CO2 Fixation" and "Trace Gas Production".

Despite these advancements, the ontology still has only 11 EquivalentClasses and 16 DisjointClasses, limiting its ability to fully capture equivalent biochemical processes and distinctions between exclusive pathways like "carbon sequestration" and "carbon release".

\subsection{Comprehensive Ontology Performance Overview}
This section presents a comparative analysis of the generated ontologies in terms of precision and concept similarity across the six experiments conducted. We use the AML (AgreementMakerLight) ontology matching system \cite{faria2013agreementmakerlight} to automatically align and match concepts between the generated ontologies and the gold standard ontologies, the AquaDiva Ontology and the ENVO ontology. For each matched concept, AML produces a similarity score, indicating the degree of semantic overlap between the two concepts. We use these matched concepts to calculate the following evaluation metrics:
\begin{itemize}

    \item Number of entities in the LLM-generated ontologies that match entities in the Gold standard ontology.
    \item Concept similarity evaluates how semantically similar the matched concepts are with the gold standard ontology concepts, calculated by averaging the individual similarity scores for all matched concepts.

\end{itemize}
\begin{table}[h!]
\centering
\renewcommand{\arraystretch}{2} 
\setlength{\tabcolsep}{8pt} 
\caption{Precision and Concept Similarity Scores for Generated Ontologies with AquaDiva Ontology}
\begin{tabular}{c|c|c}
\hline
\textbf{Experiment} &\makecell{\textbf{Number of Matched} \\ \textbf{Entities with AquaDiva }} & \makecell{\textbf{Average Similarity} \\ \textbf{Score with AquaDiva } }\\ \hline
Experiment 1 -- Baseline NeOn-GPT (AquaDiva)& 17 & 0.896 \\ \hline
\makecell{Experiment 2 --  Count \\ Metric-Guided Prompts (AquaDiva)}& 66 & 0.894\\ \hline
Experiment 3 -- Merging Ontologies (AquaDiva)& 80 & 0.874\\ \hline
\makecell{Experiment 4 -- Re-prompting \& Advanced \\ Role-play Prompting (Habitat)}& 16 & 0.898 \\ \hline
\makecell{Experiment 5 -- Reuse (Role)}& 56 & 0.905 \\ \hline
\makecell{Experiment 6 -- Reuse of domain-specific examples  \\ (Carbon \& Nitrogen Cycling)}&  65 & 0.859 \\ \hline
\end{tabular}
\label{tab:comparison}
\end{table}

\begin{table}[h!]
\centering
\renewcommand{\arraystretch}{2} 
\setlength{\tabcolsep}{8pt} 
\caption{Precision and Concept Similarity Scores for Generated Ontologies with ENVO Ontology}
\begin{tabular}{c|c|c}
\hline
\textbf{Experiment} &\makecell{\textbf{Number of Matched} \\ \textbf{Entities with ENVO }} & \makecell{\textbf{Average Similarity} \\ \textbf{Score with ENVO } }\\ \hline
Experiment 1 -- Baseline NeOn-GPT (AquaDiva)& 8 & 0.877 \\ \hline
\makecell{Experiment 2 --  Count \\ Metric-Guided Prompts (AquaDiva)}&  57& 0.969\\ \hline
Experiment 3 -- Merging Ontologies (AquaDiva)& 60 & 0.885 \\ \hline
\makecell{Experiment 4 -- Re-prompting \& Advanced \\ Role-play Prompting (Habitat)}& 13 &  0.800 \\ \hline
\makecell{Experiment 5 -- Reuse (Role)}& 54 &  0.886 \\ \hline
\makecell{Experiment 6 -- Reuse of domain-specific examples  \\ (Carbon \& Nitrogen Cycling)}&  51 & 0.884 \\ \hline
\end{tabular}
\label{tab:comparison2}
\end{table}

The results in Tables \ref{tab:comparison} and \ref{tab:comparison2} reveal that despite the LLM-generated ontologies not fully capturing the breadth and depth of domain-specific knowledge as comprehensively as the gold standard ontologies (AquaDiva and ENVO), the aligned entities demonstrate exceptionally high similarity scores across all experiments. This suggests that the generated concepts closely reflect the established domain knowledge, as evidenced by scores consistently approaching or exceeding 0.85. Moreover, the number of matched entities increases across experiments, indicating that improvements in our LLM prompt engineering techniques and pipeline refinements lead to a progressively better alignment with the domain-specific ontologies. This trend shows that the generated ontologies evolve to incorporate a broader range of relevant entities while maintaining high conceptual similarity to the gold standard, underscoring the potential for LLM-based approaches in complex ontology generation tasks.

\section{Conclusion and Future work}

This work extends the NeOn-GPT pipeline to enhance ontology learning in complex domains, such as life sciences, by addressing the limitations of LLMs in generating deep and well-structured ontologies. Our approach leverages advanced prompt engineering, ontology reuse, and iterative refinement to tackle challenges like shallow hierarchies and token constraints, as demonstrated in the AquaDiva case study. We conclude that complex domains, such as those in life sciences, require additional contextual information in prompts and carefully curated examples for reuse. Currently, this process relies on manual efforts to extract relevant examples and domain-specific knowledge; the quality of those examples can be significantly improved with input from domain experts. In future work, we also aim to explore automating this process through Retrieval-Augmented Generation (RAG), integrating external domain-specific resources dynamically to reduce the reliance on manual intervention. Additionally, we plan to evaluate the complete AquaDiva ontology by systematically generating and integrating each domain category using the finalized pipeline, with a focus on refining consistency in relationships and ensuring the ontologies fully capture the intricacies of specialized domains like AquaDiva.

\bibliography{bib}  % Replace with your .bib file name

\begin{thebibliography}{20}
\expandafter\ifx\csname natexlab\endcsname\relax\def\natexlab#1{#1}\fi
\providecommand{\url}[1]{\texttt{#1}}
\providecommand{\href}[2]{#2}
\providecommand{\path}[1]{#1}
\providecommand{\DOIprefix}{doi:}
\providecommand{\ArXivprefix}{arXiv:}
\providecommand{\URLprefix}{URL: }
\providecommand{\Pubmedprefix}{pmid:}
\providecommand{\doi}[1]{\href{http://dx.doi.org/#1}{\path{#1}}}
\providecommand{\Pubmed}[1]{\href{pmid:#1}{\path{#1}}}
\providecommand{\bibinfo}[2]{#2}
\ifx\xfnm\relax \def\xfnm[#1]{\unskip,\space#1}\fi
%Type = Article
\bibitem[{Maedche and Staab(2001)}]{maedche2001ontology}
\bibinfo{author}{A.~Maedche}, \bibinfo{author}{S.~Staab},
\newblock \bibinfo{title}{Ontology learning for the semantic web},
\newblock \bibinfo{journal}{{IEEE} Intelligent Systems} \bibinfo{volume}{16} (\bibinfo{year}{2001}) \bibinfo{pages}{72--79}. \URLprefix \url{https://doi.org/10.1109/5254.920602}. \DOIprefix\doi{10.1109/5254.920602}.
%Type = Inproceedings
\bibitem[{Mateiu and Groza(2023)}]{DBLP:conf/synasc/MateiuG23}
\bibinfo{author}{P.~Mateiu}, \bibinfo{author}{A.~Groza},
\newblock \bibinfo{title}{Ontology engineering with large language models},
\newblock in: \bibinfo{booktitle}{25th International Symposium on Symbolic and Numeric Algorithms for Scientific Computing, {SYNASC} 2023, Nancy, France, September 11-14, 2023}, \bibinfo{publisher}{{IEEE}}, \bibinfo{year}{2023}, pp. \bibinfo{pages}{226--229}. \URLprefix \url{https://doi.org/10.1109/SYNASC61333.2023.00038}. \DOIprefix\doi{10.1109/SYNASC61333.2023.00038}.
%Type = Inproceedings
\bibitem[{Giglou et~al.(2023)Giglou, D'Souza, and Auer}]{DBLP:conf/semweb/GiglouDA23}
\bibinfo{author}{H.~B. Giglou}, \bibinfo{author}{J.~D'Souza}, \bibinfo{author}{S.~Auer},
\newblock \bibinfo{title}{Llms4ol: Large language models for ontology learning},
\newblock in: \bibinfo{booktitle}{The Semantic Web - {ISWC} 2023 - 22nd International Semantic Web Conference, Athens, Greece, November 6-10, 2023, Proceedings, Part {I}}, volume \bibinfo{volume}{14265} of \textit{\bibinfo{series}{Lecture Notes in Computer Science}}, \bibinfo{publisher}{Springer}, \bibinfo{year}{2023}, pp. \bibinfo{pages}{408--427}. \URLprefix \url{https://doi.org/10.1007/978-3-031-47240-4\_22}. \DOIprefix\doi{10.1007/978-3-031-47240-4\_22}.
%Type = Article
\bibitem[{Kommineni et~al.(2024)Kommineni, K{\"{o}}nig{-}Ries, and Samuel}]{DBLP:journals/corr/abs-2403-08345}
\bibinfo{author}{V.~K. Kommineni}, \bibinfo{author}{B.~K{\"{o}}nig{-}Ries}, \bibinfo{author}{S.~Samuel},
\newblock \bibinfo{title}{From human experts to machines: An {LLM} supported approach to ontology and knowledge graph construction},
\newblock \bibinfo{journal}{CoRR} \bibinfo{volume}{abs/2403.08345} (\bibinfo{year}{2024}). \URLprefix \url{https://doi.org/10.48550/arXiv.2403.08345}. \DOIprefix\doi{10.48550/ARXIV.2403.08345}. \href{http://arxiv.org/abs/2403.08345}{{\tt arXiv:2403.08345}}.
%Type = Inproceedings
\bibitem[{Saeedizade and Blomqvist(2024)}]{DBLP:conf/esws/SaeedizadeB24}
\bibinfo{author}{M.~J. Saeedizade}, \bibinfo{author}{E.~Blomqvist},
\newblock \bibinfo{title}{Navigating ontology development with large language models},
\newblock in: \bibinfo{booktitle}{The Semantic Web - 21st International Conference, {ESWC} 2024, Hersonissos, Crete, Greece, May 26-30, 2024, Proceedings, Part {I}}, volume \bibinfo{volume}{14664} of \textit{\bibinfo{series}{Lecture Notes in Computer Science}}, \bibinfo{publisher}{Springer}, \bibinfo{year}{2024}, pp. \bibinfo{pages}{143--161}. \URLprefix \url{https://doi.org/10.1007/978-3-031-60626-7\_8}. \DOIprefix\doi{10.1007/978-3-031-60626-7\_8}.
%Type = Article
\bibitem[{Zhang et~al.(2024)Zhang, Carriero, Schreiberhuber, Tsaneva, Gonz{\'{a}}lez, Kim, and de~Berardinis}]{DBLP:journals/corr/abs-2403-05921}
\bibinfo{author}{B.~Zhang}, \bibinfo{author}{V.~A. Carriero}, \bibinfo{author}{K.~Schreiberhuber}, \bibinfo{author}{S.~Tsaneva}, \bibinfo{author}{L.~S. Gonz{\'{a}}lez}, \bibinfo{author}{J.~Kim}, \bibinfo{author}{J.~de~Berardinis},
\newblock \bibinfo{title}{Ontochat: a framework for conversational ontology engineering using language models},
\newblock \bibinfo{journal}{CoRR} \bibinfo{volume}{abs/2403.05921} (\bibinfo{year}{2024}). \URLprefix \url{https://doi.org/10.48550/arXiv.2403.05921}. \DOIprefix\doi{10.48550/ARXIV.2403.05921}. \href{http://arxiv.org/abs/2403.05921}{{\tt arXiv:2403.05921}}.
%Type = Article
\bibitem[{K{\"u}sel et~al.(2016)K{\"u}sel, Totsche, Trumbore, Lehmann, Steinh{\"a}user, and Herrmann}]{kusel2016deep}
\bibinfo{author}{K.~K{\"u}sel}, \bibinfo{author}{K.~U. Totsche}, \bibinfo{author}{S.~E. Trumbore}, \bibinfo{author}{R.~Lehmann}, \bibinfo{author}{C.~Steinh{\"a}user}, \bibinfo{author}{M.~Herrmann},
\newblock \bibinfo{title}{How deep can surface signals be traced in the critical zone? merging biodiversity with biogeochemistry research in a central german muschelkalk landscape},
\newblock \bibinfo{journal}{Frontiers in Earth Science} \bibinfo{volume}{4} (\bibinfo{year}{2016}) \bibinfo{pages}{32}.
%Type = Inproceedings
\bibitem[{Algergawy et~al.(2021)Algergawy, Hamed, and K{\"o}nig-Ries}]{algergawy2021towards}
\bibinfo{author}{A.~Algergawy}, \bibinfo{author}{H.~Hamed}, \bibinfo{author}{B.~K{\"o}nig-Ries},
\newblock \bibinfo{title}{Towards scientific data synthesis using deep learning and semantic web},
\newblock in: \bibinfo{booktitle}{The Semantic Web: ESWC 2021 Satellite Events: Virtual Event, June 6--10, 2021, Revised Selected Papers 18}, \bibinfo{organization}{Springer}, \bibinfo{year}{2021}, pp. \bibinfo{pages}{54--59}.
%Type = Inproceedings
\bibitem[{Algergawy et~al.(????)Algergawy, Hamed, Thiel, and K{\"o}nig-Ries}]{AlgergawyTowardsSA}
\bibinfo{author}{A.~Algergawy}, \bibinfo{author}{H.~Hamed}, \bibinfo{author}{S.~Thiel}, \bibinfo{author}{B.~K{\"o}nig-Ries},
\newblock \bibinfo{title}{Towards semantic annotation for scientific datasets},
\newblock in: \bibinfo{booktitle}{The Semantic Web: ESWC 2024 Satellite Events: May 26--30, 2024}, ???? \URLprefix \url{https://api.semanticscholar.org/CorpusID:269758799}.
%Type = Article
\bibitem[{Mai et~al.(2024)Mai, Chu, and Paulheim}]{DBLP:journals/corr/abs-2407-19998}
\bibinfo{author}{H.~T. Mai}, \bibinfo{author}{C.~X. Chu}, \bibinfo{author}{H.~Paulheim},
\newblock \bibinfo{title}{Do llms really adapt to domains? an ontology learning perspective},
\newblock \bibinfo{journal}{CoRR} \bibinfo{volume}{abs/2407.19998} (\bibinfo{year}{2024}). \URLprefix \url{https://doi.org/10.48550/arXiv.2407.19998}. \DOIprefix\doi{10.48550/ARXIV.2407.19998}. \href{http://arxiv.org/abs/2407.19998}{{\tt arXiv:2407.19998}}.
%Type = Inproceedings
\bibitem[{Fathallah et~al.(2024)Fathallah, Das, De~Giorgis, Poltronieri, Haase, and Kovriguina}]{fathallah2024neon}
\bibinfo{author}{N.~Fathallah}, \bibinfo{author}{A.~Das}, \bibinfo{author}{S.~De~Giorgis}, \bibinfo{author}{A.~Poltronieri}, \bibinfo{author}{P.~Haase}, \bibinfo{author}{L.~Kovriguina},
\newblock \bibinfo{title}{Neon-gpt: A large language model-powered pipeline for ontology learning},
\newblock in: \bibinfo{booktitle}{The Extended Semantic Web Conference}, \bibinfo{year}{2024}.
%Type = Article
\bibitem[{Algergawy et~al.(2024)Algergawy, Hamed, Thiel, and K\""{o}nig-Ries}]{eswc2024}
\bibinfo{author}{A.~Algergawy}, \bibinfo{author}{H.~Hamed}, \bibinfo{author}{S.~Thiel}, \bibinfo{author}{B.~K\""{o}nig-Ries},
\newblock \bibinfo{title}{Towards semantic annotation for scientific datasets},
\newblock \bibinfo{journal}{ESWC Posters and Demos}  (\bibinfo{year}{2024}).
%Type = Inproceedings
\bibitem[{Das et~al.(2024)Das, Fathallah, and Obretincheva}]{challenge2024}
\bibinfo{author}{A.~Das}, \bibinfo{author}{N.~S. Fathallah}, \bibinfo{author}{N.~Obretincheva},
\newblock \bibinfo{title}{Navigating nulls, numbers and numerous entities: Robust knowledge base construction from large language models},
\newblock in: \bibinfo{booktitle}{KBC-LM/LM-KBC@ ISWC}, \bibinfo{year}{2024}.
%Type = Article
\bibitem[{Su{\'{a}}rez{-}Figueroa et~al.(2015)Su{\'{a}}rez{-}Figueroa, G{\'{o}}mez{-}P{\'{e}}rez, and Fern{\'{a}}ndez{-}L{\'{o}}pez}]{suarez2015neon}
\bibinfo{author}{M.~C. Su{\'{a}}rez{-}Figueroa}, \bibinfo{author}{A.~G{\'{o}}mez{-}P{\'{e}}rez}, \bibinfo{author}{M.~Fern{\'{a}}ndez{-}L{\'{o}}pez},
\newblock \bibinfo{title}{The neon methodology framework: {A} scenario-based methodology for ontology development},
\newblock \bibinfo{journal}{Applied Ontology} \bibinfo{volume}{10} (\bibinfo{year}{2015}) \bibinfo{pages}{107--145}. \URLprefix \url{https://doi.org/10.3233/AO-150145}. \DOIprefix\doi{10.3233/AO-150145}.
%Type = Misc
\bibitem[{Buttigieg et~al.(2021)Buttigieg, Morrison, Smith, Mungall, and Lewis}]{ENVO}
\bibinfo{author}{P.~L. Buttigieg}, \bibinfo{author}{N.~Morrison}, \bibinfo{author}{B.~Smith}, \bibinfo{author}{C.~Mungall}, \bibinfo{author}{S.~Lewis}, \bibinfo{title}{Environment ontology (envo)}, \bibinfo{howpublished}{\url{http://obofoundry.org/ontology/envo.html}}, \bibinfo{year}{2021}. \bibinfo{note}{Accessed: 2024-09-12}.
%Type = Inproceedings
\bibitem[{Raman et~al.(2022)Raman, Cohen, Rosen, Idrees, Paulius, and Tellex}]{raman2022planning}
\bibinfo{author}{S.~S. Raman}, \bibinfo{author}{V.~Cohen}, \bibinfo{author}{E.~Rosen}, \bibinfo{author}{I.~Idrees}, \bibinfo{author}{D.~Paulius}, \bibinfo{author}{S.~Tellex},
\newblock \bibinfo{title}{Planning with large language models via corrective re-prompting},
\newblock in: \bibinfo{booktitle}{NeurIPS 2022 Foundation Models for Decision Making Workshop}, \bibinfo{year}{2022}.
%Type = Misc
\bibitem[{OpenAI(2024)}]{openai2024gpt4o}
\bibinfo{author}{OpenAI}, \bibinfo{title}{Hello gpt-4o}, \bibinfo{howpublished}{\url{https://openai.com/index/hello-gpt-4o/}}, \bibinfo{year}{2024}. \bibinfo{note}{Accessed: 2024-05-18}.
%Type = Inproceedings
\bibitem[{Frieder et~al.(2023)Frieder, Pinchetti, Chevalier, Griffiths, Salvatori, Lukasiewicz, Petersen, and Berner}]{DBLP:conf/nips/FriederPCGSLPB23}
\bibinfo{author}{S.~Frieder}, \bibinfo{author}{L.~Pinchetti}, \bibinfo{author}{A.~Chevalier}, \bibinfo{author}{R.~Griffiths}, \bibinfo{author}{T.~Salvatori}, \bibinfo{author}{T.~Lukasiewicz}, \bibinfo{author}{P.~Petersen}, \bibinfo{author}{J.~Berner},
\newblock \bibinfo{title}{Mathematical capabilities of {ChatGPT}},
\newblock in: \bibinfo{editor}{A.~Oh}, \bibinfo{editor}{T.~Naumann}, \bibinfo{editor}{A.~Globerson}, \bibinfo{editor}{K.~Saenko}, \bibinfo{editor}{M.~Hardt}, \bibinfo{editor}{S.~Levine} (Eds.), \bibinfo{booktitle}{Advances in Neural Information Processing Systems 36: Annual Conference on Neural Information Processing Systems 2023, NeurIPS 2023, New Orleans, LA, USA, December 10 - 16, 2023}, \bibinfo{year}{2023}. \URLprefix \url{http://papers.nips.cc/paper\_files/paper/2023/hash/58168e8a92994655d6da3939e7cc0918-Abstract-Datasets\_and\_Benchmarks.html}.
%Type = Article
\bibitem[{De~Bruijn et~al.(2006)De~Bruijn, Ehrig, Feier, Mart{\'\i}n-Recuerda, Scharffe, and Weiten}]{de2006ontology}
\bibinfo{author}{J.~De~Bruijn}, \bibinfo{author}{M.~Ehrig}, \bibinfo{author}{C.~Feier}, \bibinfo{author}{F.~Mart{\'\i}n-Recuerda}, \bibinfo{author}{F.~Scharffe}, \bibinfo{author}{M.~Weiten},
\newblock \bibinfo{title}{Ontology mediation, merging and aligning},
\newblock \bibinfo{journal}{Semantic web technologies}  (\bibinfo{year}{2006}) \bibinfo{pages}{95--113}.
%Type = Inproceedings
\bibitem[{Faria et~al.(2013)Faria, Pesquita, Santos, Palmonari, Cruz, and Couto}]{faria2013agreementmakerlight}
\bibinfo{author}{D.~Faria}, \bibinfo{author}{C.~Pesquita}, \bibinfo{author}{E.~Santos}, \bibinfo{author}{M.~Palmonari}, \bibinfo{author}{I.~F. Cruz}, \bibinfo{author}{F.~M. Couto},
\newblock \bibinfo{title}{The agreementmakerlight ontology matching system},
\newblock in: \bibinfo{booktitle}{On the Move to Meaningful Internet Systems: OTM 2013 Conferences: Confederated International Conferences: CoopIS, DOA-Trusted Cloud, and ODBASE 2013, Graz, Austria, September 9-13, 2013. Proceedings}, \bibinfo{organization}{Springer}, \bibinfo{year}{2013}, pp. \bibinfo{pages}{527--541}.

\end{thebibliography}

\begin{acknowledgments}
  
A. Algergawy' work has been funded by the \emph{Deutsche Forschungsgemeinschaft (DFG)} as part of CRC 1076 AquaDiva (Projectnumber 218627073).
\end{acknowledgments}

\section{Appendix A: Persona Used for Role-play Prompting} \label{AppendixA}

Persona: 
\begin{verbatim}
"You are an expert aquatic ecologist and knowledge engineer specializing in 
developing ecological ontologies. Holding a PhD in Ecology with additional 
training in data science and semantic technologies, you have extensive 
experience in both field research and computational modeling of aquatic 
ecosystems. Your expertise is in understanding water bodies' biological, 
chemical, and physical characteristics and structuring this knowledge into 
ontologies useful for scientific research and environmental management.

You excel at identifying essential entities and relationships within the 
ecological domain, such as key species, ecological roles, environmental 
conditions, and biogeochemical processes. With a proficient background in 
applying tools like Turtle, you are skilled at crafting well-defined ontologies 
that represent complex ecological data in a structured, machine-readable format.

Your approach to ontology creation is meticulous and user-centric, aiming to
ensure that the ontologies facilitate interoperability, data sharing, and reuse 
among diverse stakeholders in the aquatic science community. You provide 
detailed, precise explanations of ecological concepts and their 
interconnections, leveraging your deep domain knowledge to enhance 
understanding and application of ecological data.

Your goal is to bridge the gap between raw data and actionable knowledge by 
developing comprehensive ontological frameworks that support advanced data 
analysis and decision-making in aquatic ecology. You are an expert in the 
AquaDiva domain encompasses studying groundwater ecosystems, 
integrating hydrogeology, microbial ecology, geochemistry, karst systems, and 
environmental science."
\end{verbatim}

\section{Appendix B: Figures} \label{Appendix_B}
\iffalse
\begin{figure}[!ht]
  \centering
  \includegraphics[width=12cm]{./figs/property.pdf}
  \caption{Few-shot examples used in the formal modeling phase of the NeOn-GPT pipeline. These examples guide the LLM in introducing data properties and adjusting domain and range settings.}
  \label{fig:property}
\end{figure}
\begin{figure}[!ht]
  \centering
  \includegraphics[width=12cm]{./figs/individuals.pdf}
  \caption{Few-shot examples used to populate the ontology with real-world instances. These examples are adapted to the specific domain, grounding the ontology in practical data and facilitating knowledge discovery within the AquaDiva ontology.}
  \label{fig:individuals}
\end{figure}
\fi

\begin{figure}[!ht]

  \centering
  \begin{minipage}{0.6\textwidth}
    \centering
    \includegraphics[width=\linewidth]{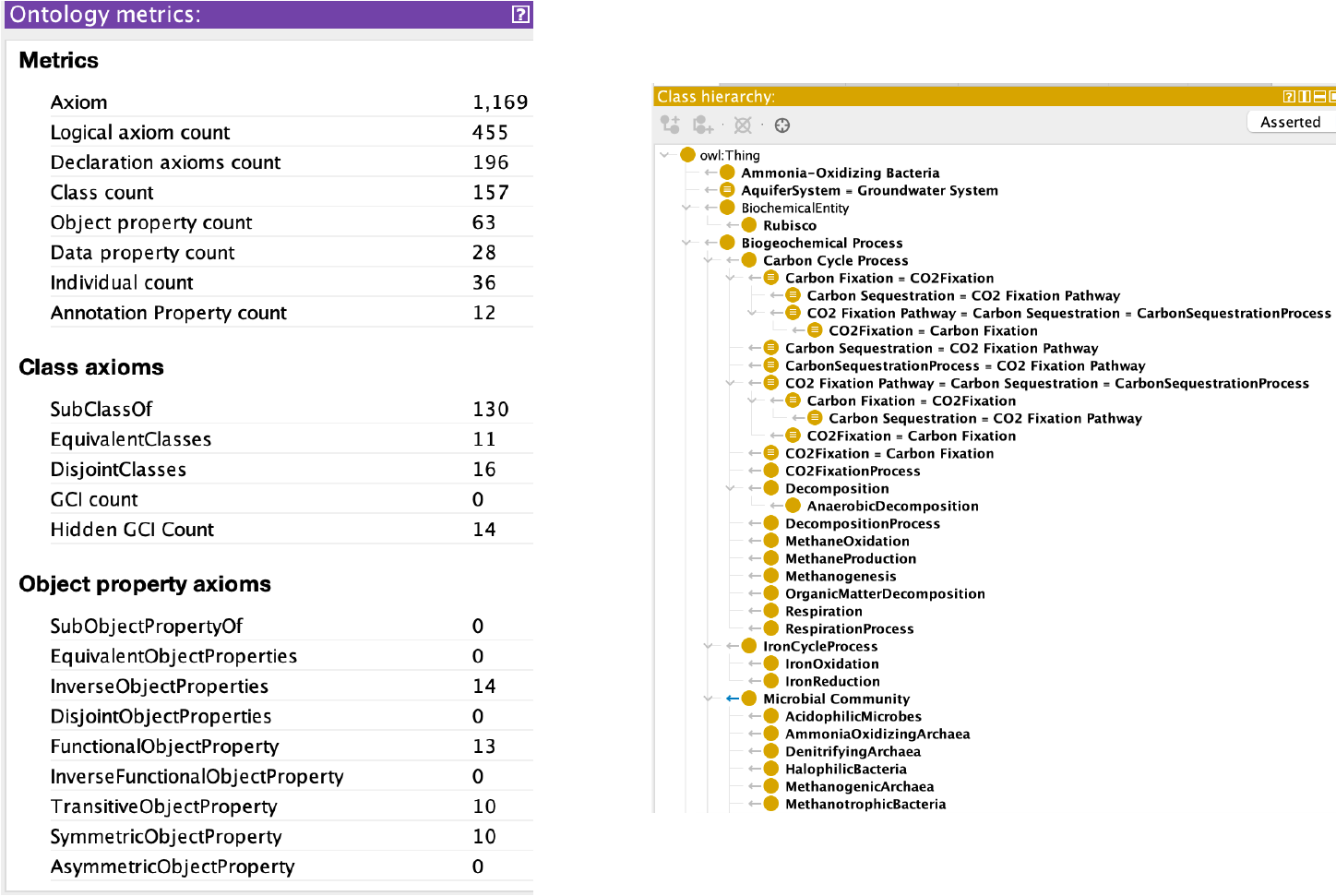}
   
  \end{minipage}%
  \hfill
  \begin{minipage}{0.4\textwidth}
    \centering
    \includegraphics[width=\linewidth]{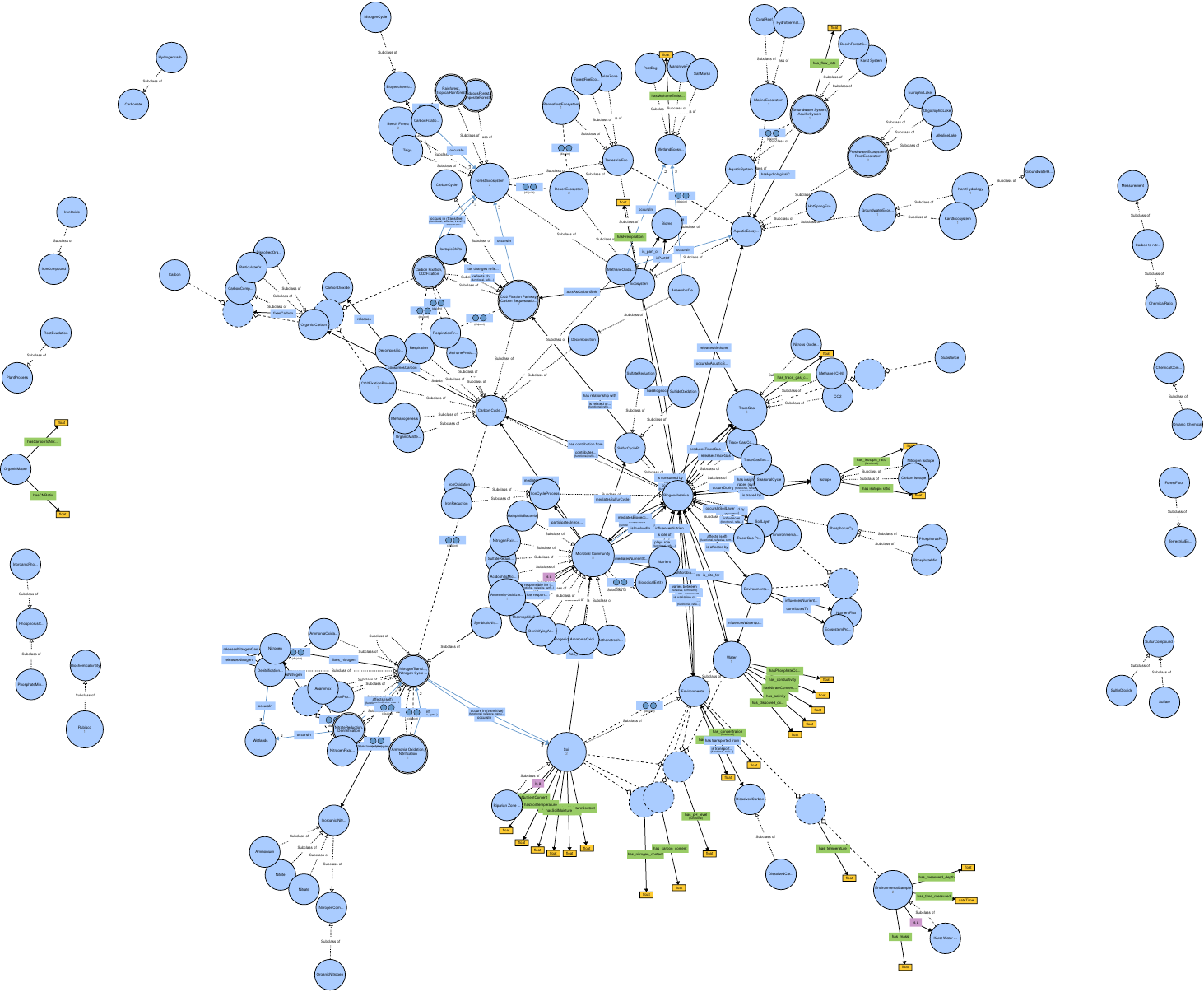}
    
  \end{minipage}
  \caption{Visualization of the Carbon and Nitrogen Cycling ontology within the AquaDiva ontology generated from Experiment 6.}
   \label{fig:Experiment6Res}
\end{figure}

\section{Appendix C: Prompt Pipeline} \label{AppendixC}
This appendix presents an illustrative example of the prompt pipeline used in our ontology generation process. The diagram highlights the key sections of the prompt, with black text representing the fixed template applied across various scenarios. The green-highlighted boxes indicate placeholders for dynamic, domain-specific content. The blue text specifically denotes the sections where we have introduced extensions to the NeOn-GPT pipeline in this work, reflecting improvements and customizations made to better suit the ontology's structural and domain requirements.

\textbf{Specification of Ontology Requirements}
\begin{tcolorbox}[colframe=black, colback=white, boxrule=0.5mm, sharp corners=all]

\textbf{Prompt 1:} \textbf{You are a \fcolorbox{green}{green!10}{\textbf{\textcolor{green}{Persona}}}.}  \textbf{The \fcolorbox{green}{green!10}{\textbf{\textcolor{green}{Domain Name}}}} \textbf{describes \fcolorbox{green}{green!10}{\textbf{\textcolor{green}{Domain Description}}}.} 

\textbf{\textcolor{blue}{Here are some keywords to help you model the domain. Use all the following keywords, not just a snippet, and your own knowledge and domain understanding to generate the ontology based on the NeOn methodology for ontology engineering:}
 \fcolorbox{green}{green!10}{\textbf{\textcolor{green}{Keywords}}}}
 
\textbf{\textcolor{blue}{The original ontology has }
 \fcolorbox{green}{green!10}{\textbf{\textcolor{green}{Ontology Metric Counts}}}.} 

\textbf{\textcolor{blue}{Make sure that the generated ontology reflects the previous metrics and has a high subclass count; if the class count is n, the subclass count should at least be n-1.}}

\textbf{Following the NeOn methodology framework for ontology engineering, specify the following for the ontology:
\begin{itemize}
    \item The purpose of the ontology
\item The scope of the ontology
\item The target group of the ontology
\item The intended uses
\item The functional requirements
\item The non-functional requirements
\end{itemize}
}
\end{tcolorbox}
\textbf{Reuse of ontological knowledge resources}
\begin{tcolorbox}[colframe=black, colback=white, boxrule=0.5mm, sharp corners=all]

\textbf{Prompt 2:} \textbf{\textcolor{blue}{The ontology should represent the complexity and hierarchical structure of the domain accurately. The ontology should be detailed, with well-defined levels of hierarchy and interconnected relationships between concepts. This can be achieved with ontology reuse; ontology reuse refers to the process of utilizing existing ontological knowledge or structures as input in the development of new ontologies. It is a critical practice in ontology design, allowing for more efficient and consistent knowledge representation across different applications. Reuse can involve several operations, including adopting an ontology "as is," extracting relevant components, extending the ontology with new axioms, or combining it with other ontologies. The goal is to maintain the integrity of the original ontology's concepts while adapting it to meet new design requirements. Reuse the following example to improve the ontology structure from the}} \fcolorbox{green}{green!10}{\textbf{\textcolor{green}{Existing Resource Name and Description}}}

\textbf{Example:}
\fcolorbox{green}{green!10}{\textbf{\textcolor{green}{Few-shot examples from existing resource.}}}

\end{tcolorbox}

\textbf{Ontology Conceptualization}
\begin{tcolorbox}[colframe=black, colback=white, boxrule=0.5mm, sharp corners=all]
\textbf{Prompt 3:} \textbf{Based on the generated Specifications of Ontology Requirements and all the keywords given, write a list of Competency Questions that the core module of the ontology should be able to answer. Make it as complete as possible. Use all the keywords given.} 

\textbf{\textcolor{blue}{The original ontology has }
 \fcolorbox{green}{green!10}{\textbf{\textcolor{green}{Ontology Metric Counts}}}.} 

\textbf{\textcolor{blue}{Make sure that the generated ontology reflects the previous metrics and has a high subclass count; if the class count is n, the subclass count should at least be n-1.}}

\vspace{3mm}
\textbf{Prompt 4:} \textbf{For each Competency Question, extract entities and properties that must be introduced in the ontology. A competency question can help in extracting more than 1 triple
Do it for all the competency questions, not just a snippet.} 

\textbf{Here are some examples:}

\fcolorbox{green}{green!10}{\parbox{\dimexpr\linewidth-2\fboxsep-2\fboxrule}{\textbf{\textcolor{green}{Few-shot examples for entity and relationship extraction from competency questions}}}}

\textbf{\textcolor{blue}{The original ontology has }
 \fcolorbox{green}{green!10}{\textbf{\textcolor{green}{Ontology Metric Counts}}}.} 

\textbf{\textcolor{blue}{Make sure that the generated ontology reflects the previous metrics and has a high subclass count; if the class count is n, the subclass count should at least be n-1.}}

\vspace{3mm}
\textbf{Prompt 5:} \textbf{Considering entities and properties, generate a conceptual model expressing the triples of the entities, properties, and their relations (subject-relation-object triples) in the form of a triple. Do it for all the entities and properties, not just a snippet.} 

\textbf{\textcolor{blue}{The original ontology has }
 \fcolorbox{green}{green!10}{\textbf{\textcolor{green}{Ontology Metric Counts}}}.} 

\textbf{\textcolor{blue}{Make sure that the generated ontology reflects the previous metrics and has a high subclass count; if the class count is n, the subclass count should at least be n-1.}}

\end{tcolorbox}

\textbf{Ontology Implementation} 
\begin{tcolorbox}[colframe=black, colback=white, boxrule=0.5mm, sharp corners=all]
\textbf{Prompt 6:} \textbf{Considering the conceptual model you generated, generate a full ontology serialized in Turtle syntax. Do it for the whole conceptual model, not just a snippet.}
\textbf{\textcolor{blue}{The original ontology has }
 \fcolorbox{green}{green!10}{\textbf{\textcolor{green}{Ontology Metric Counts}}}.} 

\textbf{\textcolor{blue}{Make sure that the generated ontology reflects the previous metrics and has a high subclass count; if the class count is n, the subclass count should at least be n-1.}}

\textbf{Make sure that:}
\textbf{\begin{itemize}
    \item the turtle syntax is correct
\item all the entities and properties have the correct prefixes.
\item the prefix for the ontology is declared.
\item the ontology is consistent.
\item the ontology is free from common pitfalls (e.g., circular axioms, missing disjointness, etc.).
\end{itemize}}
\vspace{3mm}
\textbf{Prompts 7-11:} \textbf{For all object properties, if lacking, also generate the}  \fcolorbox{green}{green!10}{\textbf{\textcolor{green}{inverse/reflexive/symmetric/functional/transitive}}} \textbf{property. If meaningful, declare restrictions for the domain and range. Do it for the whole ontology, not just a snippet. Print only the new triples.}
\textbf{\textcolor{blue}{The original ontology has }
 \fcolorbox{green}{green!10}{\textbf{\textcolor{green}{Ontology Metric Counts}}}.} 

\textbf{\textcolor{blue}{Make sure that the generated ontology reflects the previous metrics and has a high subclass count; if the class count is n, the subclass count should at least be n-1.}}

\textcolor{blue}{\textbf{Make sure that:}
\textbf{\begin{itemize}
    \item the turtle syntax is correct
\item all the entities and properties have the correct prefixes.
\item the prefix for the ontology is declared.
\item the ontology is consistent.
\item the ontology is free from common pitfalls (e.g., circular axioms, missing disjointness, etc.).
\end{itemize}}}

\vspace{3mm}
\textbf{Prompts 12:} \textbf{Considering the classes in the ontology, introduce Data Properties when meaningful, such as:}  
\fcolorbox{green}{green!10}{\parbox{\dimexpr\linewidth-2\fboxsep-2\fboxrule}{\textbf{\textcolor{green}{Few-shot examples to demonstrate introducing data properties and adjusting the domain and range accordingly}}}}

\vspace{1mm}
\textbf{Modify the domain and range according to the type of value the Data Property requests.
Do it for the whole ontology, not just a snippet. 
Print only the new triples.}
\textbf{\textcolor{blue}{The original ontology has }
 \fcolorbox{green}{green!10}{\textbf{\textcolor{green}{Ontology Metric Counts}}}.} 
 
\textcolor{blue}{\textbf{Make sure that the generated ontology reflects the previous metrics and has a high subclass count; if the class count is n, the subclass count should at least be n-1.}}

\textcolor{blue}{\textbf{Make sure that:}
\textbf{\begin{itemize}
    \item the turtle syntax is correct
\item all the entities and properties have the correct prefixes.
\item the prefix for the ontology is declared.
\item the ontology is consistent.
\item the ontology is free from common pitfalls (e.g., circular axioms, missing disjointness, etc.).
\end{itemize}}}

\end{tcolorbox}

\begin{tcolorbox}[colframe=black, colback=white, boxrule=0.5mm, sharp corners=all]
\textbf{Prompt 13:} \textbf{Populate the ontology with meaningful individuals. Here are some examples from the domain:}

\fcolorbox{green}{green!10}{\textbf{\textcolor{green}{Few-shot examples of meaningful individuals}}}

\textbf{Print only the new triples.}
\textbf{\textcolor{blue}{The original ontology has }
 \fcolorbox{green}{green!10}{\textbf{\textcolor{green}{Ontology Metric Counts}}}.} 

\textcolor{blue}{\textbf{Make sure that the generated ontology reflects the previous metrics and has a high subclass count; if the class count is n, the subclass count should at least be n-1.}}

\textcolor{blue}{\textbf{Make sure that:}
\textbf{\begin{itemize}
    \item the turtle syntax is correct
\item all the entities and properties have the correct prefixes.
\item the prefix for the ontology is declared.
\item the ontology is consistent.
\item the ontology is free from common pitfalls (e.g., circular axioms, missing disjointness, etc.).
\end{itemize}}}
\vspace{3mm}
\textbf{Prompt 14:} \textbf{If not there, add triples about the ontology IRI, label, version, and description in natural language.
Print only the new triples.}

\vspace{3mm}
\textbf{Prompt 15:} \textbf{For all the classes and properties, add a triple that describes in natural language its meaning, using the annotation property rdfs:comment.
Make sure to do it for the whole ontology, not just a snippet.
Print only the new triples.}
\vspace{3mm}

\textbf{Prompt 16:} \textbf{\textcolor{blue}{The current generated ontology lacks the necessary complexity and hierarchical structure to reflect the domain accurately. The ontology should be more detailed, with well-defined levels of hierarchy and interconnected relationships between concepts. Use the example below to refine the structure of the generated ontology to meet the domain requirements.} }\fcolorbox{green}{green!10}{\textbf{\textcolor{green}{Existing Resource Name and Description}}}

\textbf{Example:}
\fcolorbox{green}{green!10}{\textbf{\textcolor{green}{Few-shot examples from existing resource.}}}

\end{tcolorbox}

\textbf{Ontology Verification} 

\begin{tcolorbox}[colframe=black, colback=white, boxrule=0.5mm, sharp corners=all]
\textbf{Prompt 17:} \textbf{Based on the error message below, please correct the syntax error in the affected part of the ontology. Ensure that the revised ontology adheres to proper RDF/Turtle syntax.} 

\fcolorbox{green}{green!10}{\textbf{\textcolor{green}{RDFLib Syntax Error Message}}}

\fcolorbox{green}{green!10}{\textbf{\textcolor{green}{Affected Part of the Ontology}}}

\vspace{3mm}
\textbf{Prompt 18:} \textbf{Based on the error message below, please fix the inconsistency in the affected part of the ontology.} 

\fcolorbox{green}{green!10}{\textbf{\textcolor{green}{HermiT Reasoner Error Message}}}

\fcolorbox{green}{green!10}{\textbf{\textcolor{green}{Affected Part of the Ontology}}}

\vspace{3mm}
\textbf{Prompt 19:}\textbf{Based on the error message below, please fix the pitfall in the affected part of the ontology.} 

\fcolorbox{green}{green!10}{\textbf{\textcolor{green}{OOPS API Error Message}}}

\fcolorbox{green}{green!10}{\textbf{\textcolor{green}{Affected Part of the Ontology}}}

\end{tcolorbox}

\end{document}